\documentclass{article}

\usepackage[preprint]{neurips_data_2024}

\usepackage[utf8]{inputenc} 
\usepackage[T1]{fontenc}    
\usepackage{hyperref}       
\usepackage{url}            
\usepackage{booktabs}       
\usepackage{amsfonts}       
\usepackage{nicefrac}       
\usepackage{microtype}      
\usepackage{xcolor}         
\usepackage{graphicx}
\usepackage{subcaption}
\usepackage{dirtytalk}
\usepackage{arydshln}
\usepackage{xcolor,colortbl}
\usepackage{tabularx}
\usepackage{float}
\usepackage{graphicx} 
\usepackage{pgffor}   
\newcolumntype{C}{>{\centering\arraybackslash}X}
\usepackage{multirow}

\title{VLM4Bio: A Benchmark Dataset to Evaluate Pretrained Vision-Language Models for Trait Discovery from Biological Images}
\author{%
M. Maruf$^{1}$\thanks{Corresponding authors: \textit{\{marufm, karpatne\}@vt.edu}} \quad Arka Daw$^{9*}$ \quad Kazi Sajeed Mehrab$^1$ \quad Harish Babu Manogaran$^1$ \\ \textbf{Abhilash Neog}$^1$ \quad \textbf{Medha Sawhney}$^1$ \quad \textbf{Mridul Khurana}$^1$ \quad  
\textbf{James P. Balhoff}$^3$ \quad \\ \textbf{Yasin Bakış}$^4$ \quad \textbf{Bahadir Altintas}$^4$ \quad 
\textbf{Matthew J Thompson}$^5$ \quad
\textbf{Elizabeth G Campolongo}$^5$  \\  \textbf{Josef C. Uyeda}$^1$ \quad \textbf{Hilmar Lapp}$^6$ \quad \textbf{Henry L. Bart Jr.}$^4$ \quad \textbf{Paula M. Mabee}$^7$ \\ \textbf{Yu Su}$^5$ \quad \textbf{Wei-Lun Chao}$^5$ \quad \textbf{Charles Stewart}$^8$ \quad \textbf{Tanya Berger-Wolf}$^5$ \\ 
\textbf{Wasila Dahdul}$^2$ \quad \textbf{Anuj Karpatne}$^{1*}$ \\
$^1$Virginia Tech \quad $^2$Univ. of California, Irvine \quad $^3$UNC at Chapel Hill \quad $^4$Tulane Univ. \\ $^5$Ohio State Univ. \: $^6$Duke Univ. \: $^7$Battelle \: $^8$Rensselaer Polytechnic Institute \: $^9$Oak Ridge Lab.\\
}

\begin{document}

\maketitle
\everypar{\looseness=-1}

\begin{figure}[h]
  \centering
  \includegraphics[width=\textwidth]{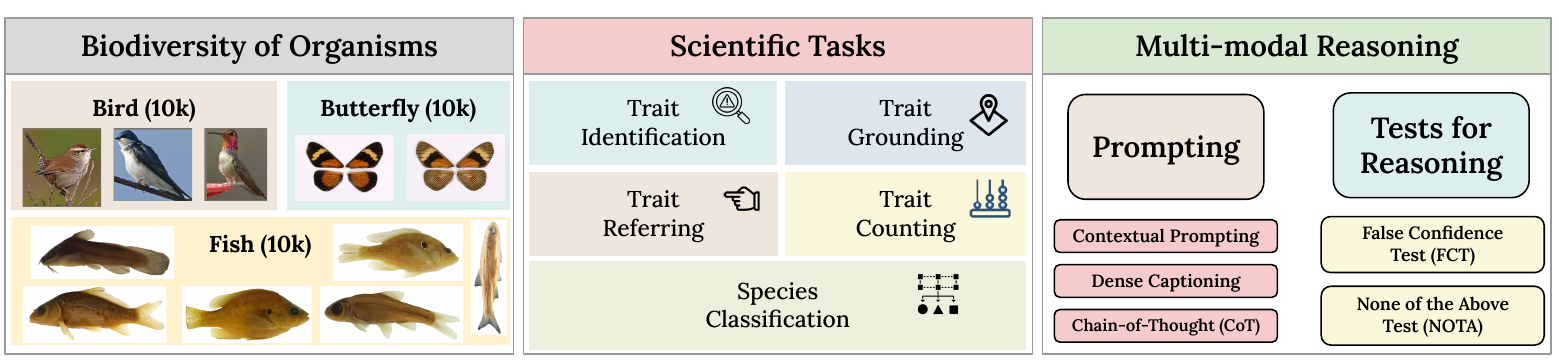}
  \caption{
  \small{Overview of our goals and contributions. We analyze the capabilities of 12 state-of-the-art (SOTA) vision-language models (VLMs) in answering scientific questions using images from three groups of organisms: fishes, birds, and butterflies, over five groups of biologically relevant tasks. We also explore the effectiveness of these models for reasoning using various prompting techniques and tests for reasoning hallucination.}}
  \label{fig:teaser}
\end{figure}

\begin{abstract}
  Images are increasingly becoming the currency for documenting biodiversity on the planet, 
  providing novel opportunities for accelerating scientific discoveries in the field of organismal biology, especially
  with the advent of large vision-language models (VLMs). We ask if pre-trained VLMs can aid scientists in answering a range of biologically relevant questions without any additional fine-tuning. 
  In this paper, we evaluate the effectiveness of 12 state-of-the-art (SOTA) VLMs in the field of organismal biology using a novel dataset, \textbf{VLM4Bio}, consisting of $~469K$ question-answer pairs involving $30K$ images from three groups of organisms: fishes, birds, and butterflies, covering five biologically relevant tasks. We also explore the effects of applying prompting techniques and tests for reasoning hallucination on the performance of VLMs, shedding new light on the capabilities of current SOTA VLMs in answering biologically relevant questions using images.
  \footnote{The code and datasets for running all the analyses reported in this paper can be found at \url{https://github.com/sammarfy/VLM4Bio}.}
  
\end{abstract}

%
%
\addtocontents{toc}{\protect\setcounter{tocdepth}{0}}

\section{Introduction}
\vspace{-2ex}
There is a growing deluge of images that are being collected, stored, and shared in organismal biology---the branch of biology interested in the study of structure, ecology, and evolution of organisms. In particular, images are increasingly becoming the currency for documenting the vast array of biodiverse organisms on our planet, with repositories containing millions of images of biological specimens collected by scientists in field museums or captured by drones, camera traps, or tourists posting photos on social media. 
This growing wealth of biological images provides a unique opportunity to 
understand
the scientific mechanisms of how organisms evolve and adapt to their environment directly from images. The traditional approach for advancing  knowledge in 
organismal biology is by discovering the observable characteristics of organisms or \textit{traits} (e.g., {beak color, stripe pattern, and fin curvature}) that serve a variety of biological tasks such as defining groups of organisms, understanding their genetic and developmental underpinnings, and analyzing their interactions with environmental selection pressures \cite{houlerossoni2022}. However, the measurement of traits is not straightforward and often relies on expert visual attention involving labor-intensive operations and subjective definitions \cite{simoes2017giant}, hindering rapid scientific advancement \cite{lurig2021}. 

With the recent rise of large foundation models such as vision-language models (VLMs) (e.g., GPT-4, GPT-4V(ision) \cite{gpt4, gpt4v}, Gemini \cite{team2023gemini}, LLaMA, and LLaVA \cite{touvron2023llama, liu2023improved}) that can simultaneously solve a diverse range of tasks involving text and images, it is pertinent to ask if pre-trained VLMs 
contain the necessary \textit{scientific knowledge} 
to aid biologists in answering a variety of questions pertinent to the discovery of biological traits from images. 
Note that unlike mainstream tasks in computer vision, understanding scientific images requires knowledge of domain-specific terminologies and reasoning capabilities that are not fully represented in conventional image datasets used for training VLMs. In particular, an important end-goal in scientific applications such as organismal biology is to explain the process of visual reasoning used to arrive at a prediction, often involving the knowledge of biological traits. Hence, to assess the usefulness of VLMs in accelerating discoveries in organismal biology, it is important to test their ability to identify and reason about biological traits automatically from images.

In this work, we assess the zero-shot capabilities of 12 state-of-the-art (SOTA) VLMs, including the proprietary GPT-4V(ision) and the recent GPT-4o along with other open-source VLMs, on five scientifically relevant tasks in organismal biology, namely species classification, trait identification, trait grounding, trait referring, and trait counting. These tasks are designed to test different facets of VLM performance in organismal biology, ranging from measuring predictive accuracy to assessing their ability to reason about their predictions using visual cues of known biological traits. For example, the task of species classification tests the ability of VLMs to discriminate between species, while in trait grounding and referring, we specifically test if VLMs are able to localize morphological traits (e.g., the presence of fins or patterns and colors of birds) within the image. 
To perform this evaluation, we present \textbf{VLM4Bio}, a benchmark dataset of $\approx 469K$ question-answer pairs based on $30k$ images of three taxonomic groups of organisms: fishes, birds, and butterflies.

\par \noindent \textbf{Contributions:} (1) We present a novel dataset of scientific question-answer pairs to evaluate the effectiveness of VLMs in answering scientific questions across a range of biologically relevant tasks in the field of organismal biology. (2) We present novel benchmarking analyses of the zero-shot effectiveness of pre-trained SOTA VLMs on our dataset, exposing their gaps in advancing scientific knowledge of organismal biology. (3) We present novel comparisons studying the effects of prompting and tests for reasoning hallucination on VLM performance, shedding new light on the reasoning capabilities of SOTA VLMs in organismal biology.

\begin{figure}[t]
    \centering
    \includegraphics[width=0.8\linewidth]{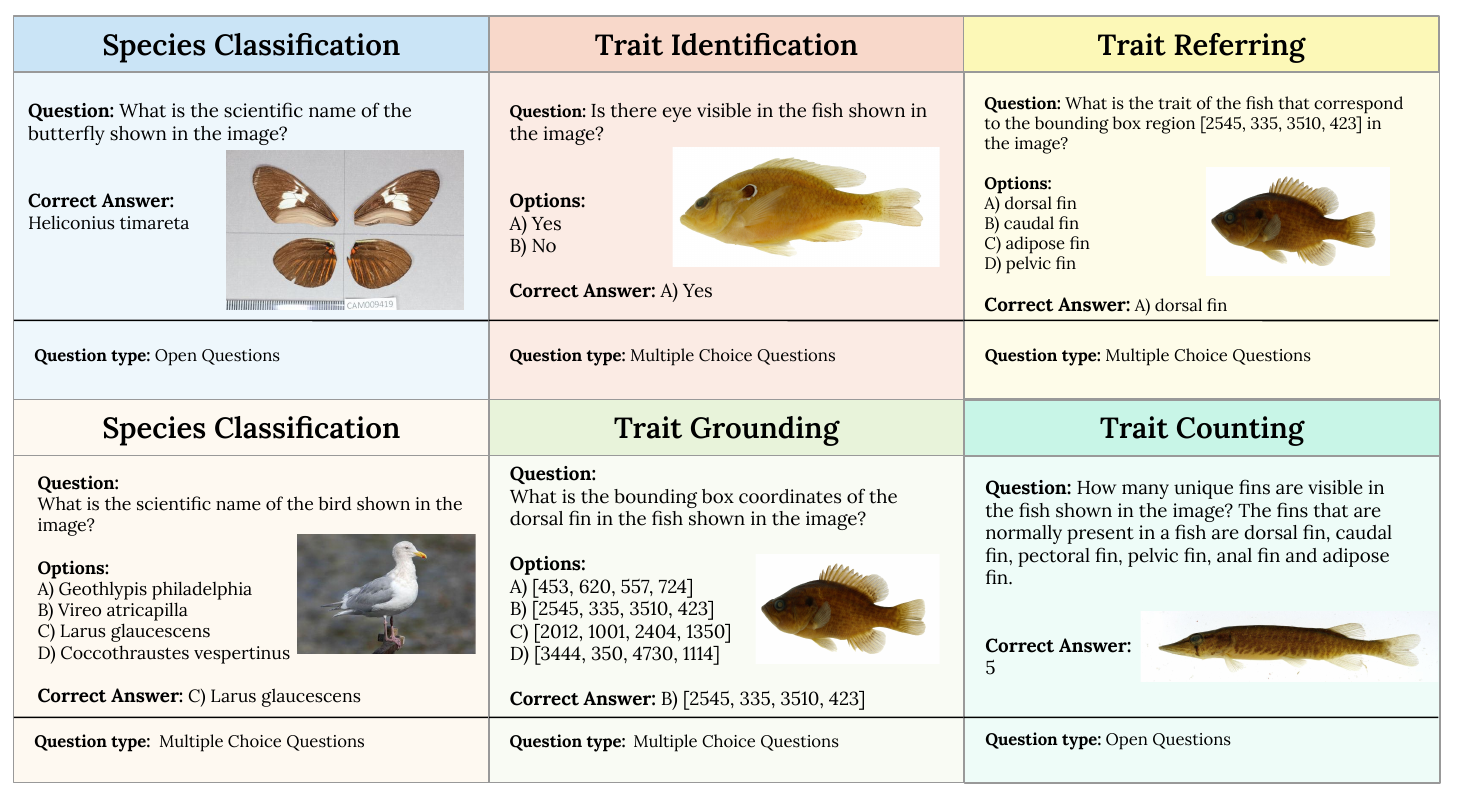}
    \caption{Illustrative examples of \textbf{VLM4Bio} tasks with different question-types.}
    \label{fig:sci-task}
    
\end{figure}
\section{Related Works}
With the rise of SOTA VLMs such as GPT-4V(vision) \cite{gpt4v}, GPT-4o \cite{gpt4o}, and Gemini \cite{team2023gemini}, there has been a simultaneous growth in the number of benchmarking analyses published in the last few years to evaluate different facets of VLM performance on a range of mainstream tasks in computer vision. A majority of previous analyses \cite{yang2022empirical,yue2023mmmu} involve evaluations on single tasks like Visual Question Answering (VQA), OK-VQA \cite{marino2019ok}, MSCOCO \cite{lin2014microsoft}, and GQA \cite{hudson2019gqa}. Other datasets such as POPE \cite{li2023evaluating}, HaELM \cite{touvron2023llama}, LAMM \cite{yin2023lamm}, MMBench \cite{liu2023mmbench}, MM-Vet \cite{yu2023mm}, LVLM-eHub \cite{xu2023lvlm}, SEED \cite{li2023seed}, and GAIA \cite{mialon2023gaia} have also been developed to evaluate the capabilities of VLMs on complex tasks such as reasoning and ability to handle multimodal data. There are also some recent domain-specific benchmark datasets, such as MathVista \cite{lu2023mathvista}, which includes a variety of challenging VQA problems in the mathematical domain, MedQA(USMLE) \cite{jin2021disease} which is a collection of VQA problems from medical exams, and the recent MMMU \cite{yue2023mmmu} dataset, which covers expert-level problems from diverse fields such as business, arts, science, health, medicine, and engineering. 

VLM4Bio dataset is different from existing benchmarks involving domain-specific datasets because of the following reasons. (1) \textit{Focus on organismal biology}: While previous works have focused on benchmarking the performance of VLMs on other scientific domains (e.g., Arts and Design, Business, Health, and Medicine in MMMU \cite{yue2023mmmu} or Mathematics in MathVista \cite{lu2023mathvista}), there exists no previous VQA benchmark dataset in the domain of organismal biology to the best of our knowledge. Our work thus fills a critical gap in evaluating the performance of VLMs in a field of biology that has several societal implications such as monitoring biodiversity and understanding the impact of climate change on species traits and populations. (2) \textit{Breadth of Evaluation Tasks}: While previous works are tailored to one or a few evaluation tasks, we consider a wide range of tasks motivated by the needs of domain scientists in the field of organismal biology. They include predictive tasks such as species classification and trait identification as well as tasks that require visual reasoning including trait grounding and referring. We also provide novel comparisons about the performance of VLMs on both open-ended and multiple-choice question (MCQ) formats and comparisons over predictive as well as visual reasoning tasks, in contrast to prior works. 

\section{VLM4Bio Tasks}

Figure \ref{fig:sci-task} shows illustrative examples of the five VLM4Bio tasks relevant to biologists that we consider in our study, described in detail in the following. 

\par \textbf{Species Classification:} A common (and often the first) task that a biologist considers when examining an organism specimen is to identify its scientific name (or species class). Hence,  we consider asking a VLM to provide the scientific name of the organism shown in a given image. There are two types of questions that we consider for this task. First, we consider \textit{open-ended questions}, where we do not provide any answer choices (or options) to the VLM in the input prompt. The second type is \textit{multiple-choice (MC) questions}, where we provide four choices of candidate species names for the VLM to choose from (out of which only one is correct while the remaining three are randomly selected from the set of all species classes).

\par \textbf{Trait Identification:} An important goal in organismal biology is to answer questions regarding the observable characteristics of organisms, also known as traits. 
We thus consider asking VLMs to identify a particular trait of an organism given its image for two taxonomic groups: fishes and birds. For fishes, we considered 10 binary (presence/absence) traits and generated MC questions for the presence of each trait in an image (with two options: yes or no), whereas for birds, we considered 28 traits covering their color, pattern, and measurements (size and shape of regions) in a multiple-choice format. We provide a detailed list of all fish and bird traits in the Supplementary. 

\par \textbf{Trait Grounding and Referring:}
To further understand the ability of VLMs to visually explain the reasoning behind their prediction of a trait, it is important to evaluate if a VLM correctly identifies the region in the image containing the trait. For this purpose, we consider two other tasks: trait grounding \& trait referring, for the taxonomic groups of fishes and birds. 
In the first task of trait grounding, we ask the VLM to locate a given trait of an organism on its image (i.e., \textit{text to location}). We consider MC question-format for this task where we provide four options of bounding boxes in the image as candidate answer choices, where one of the bounding boxes correctly contains the trait while the remaining three are randomly sampled from the set of bounding boxes containing other traits of the organism. In the second task of trait referring, we consider the opposite scenario where we provide a bounding box as input to the VLM and ask it to identify the name of the trait present in the bounding box (i.e., \textit{location to text}). We again provide four answer choices in MC question-format, where only one of the options is correct while the remaining three are randomly sampled from the names of other traits of the organism. 

\par \textbf{Trait Counting:} We simply ask how many traits are present in an image of a fish specimen. This is biologically relevant, for example, to understand the number of fins present in a fish organism. Similar to the species classification task, we have open and MC question-types for this task.




\section{VLM4Bio Dataset}

\par \noindent \textbf{Data Collection and Preprocessing}: 
We collected images of three taxonomic groups of organisms: fish, birds, and butterflies, each containing around $10K$ images. 
Images for fish (\textbf{Fish-10K}) were curated from the larger image collection, FishAIR \cite{fishair}, which contains images from the Great Lakes Invasive Network Project (GLIN) \cite{GLIN} and Integrated Digitized Biocollections (iDigBio) \cite{iDigBio}. These images originate from various museum collections such as INHS \cite{inhs}, FMNH \cite{fmnh}, OSUM \cite{osum}, JFBM \cite{JFBM}, UMMZ \cite{UMMZ} and UWZM \cite{uwzm}.
We created the Fish-10K dataset by randomly sampling $10K$ images and preprocessing the images to crop and remove the background.
For consistency, we leverage GroundingDINO \cite{liu2023grounding} to crop the fish body from the background and Segment Anything Model (SAM) \cite{kirillov2023segment} to remove the background.
We curated the images for butterflies (\textbf{Butterfly-10K}) from the Jiggins Heliconius Collection dataset \cite{lawrence_campolongo_j2024}, which has images collected from various sources \footnote{Sources: \cite{but1_gabriela_montejo_kovacevich_2020_4289223, but2_patricio_a_salazar_2020_4288311, but3_montejo_kovacevich_2019_2677821, but4_jiggins_2019_2682458, but5_montejo_kovacevich_2019_2682669, but6_montejo_kovacevich_2019_2684906, but7_warren_2019_2552371, but8_warren_2019_2553977, but9_montejo_kovacevich_2019_2686762, but10_jiggins_2019_2549524, but11_jiggins_2019_2550097, but12_joana_i_meier_2020_4153502, but13_montejo_kovacevich_2019_3082688, but14_montejo_kovacevich_2019_2813153, but15_salazar_2018_1748277, but16_montejo_kovacevich_2019_2702457, but17_salazar_2019_2548678, but18_pinheiro_de_castro_2022_5561246, but19_montejo_kovacevich_2019_2707828, but20_montejo_kovacevich_2019_2714333, but21_gabriela_montejo_kovacevich_2020_4291095, but22_montejo_kovacevich_2021_5526257, but23_warren_2019_2553501, but24_salazar_2019_2735056, but25_mattila_2019_2555086}}. We carefully sampled $10K$ images for Butterfly-10K from the entire collection to ensure the images capture unique specimens and represent a diverse set of species by adopting the following two steps. First, we filter out images with more than one image from the same view (i.e., dorsal or ventral).
Second, we ensure each species has a minimum of $20$ images and no more than $2,000$ images. The images for birds (\textbf{Bird-10K}) are obtained from the CUB-200-2011 \cite{wah2011caltech} dataset by taking 190 species for which the common name to scientific name mapping is available. This results in a fairly balanced dataset with around $11K$ images in total. Additional details on dataset preprocessing are provided in the Supplementary.
 
\begin{table*}[t]
\centering
\begin{tabular}{crrrrr}
\toprule
Statistics & \textbf{Fish-10K} & \textbf{Bird-10K} & \textbf{Butterfly-10K} & \textbf{Fish-500} & \textbf{Bird-500}\\
\midrule
\# Images  & 10,347 & 11,092 & 10,013 & 500 & 492  \\
\# Species & 495    & 188    & 60     & 60    & 47     \\
\# Genera  & 178    & 114    & 27    & 18    & 33     \\
\# Traits  & 10     & 28     & -     & 8     & 5      \\
\bottomrule
\end{tabular}
\caption{Key statistics of the \textbf{VLM4Bio} dataset.}
\label{tab:dataset-stat-main}
\end{table*}

\par \noindent \textbf{Annotation:} The scientific names for the images of Fish-10K and Butterfly-10K were obtained directly from their respective sources. For Bird-10K, we obtained the scientific names from the iNatLoc500 \cite{iNatLoc500} dataset. We curated around  $31K$ question-answer pairs in both open and multiple-choice (MC) question-formats for evaluating species classification tasks. The species-level trait presence/absence matrix for Fish-10K was manually curated with the help of biological experts co-authored in this paper. We leveraged the Phenoscape knowledge \cite{edmunds2015phenoscape} base with manual annotations to procure the presence-absence trait matrix.
For Bird-10K, we obtained the trait matrix from the attribute annotations provided along with CUB-200-2011. We constructed approximately  $380K$ question-answer pairs for trait identification tasks. 
For grounding and referring VQA tasks, the ground truths were manually annotated with the help of expert biologists on our team. We manually annotated bounding boxes corresponding to the traits of 500 fish specimens and 500 bird specimens, which are subsets of the larger Fish-10K and Bird-10K datasets, respectively. In particular, we considered  $8$ fish traits and $5$ bird traits for annotating their bounding boxes, resulting in a total of  $26K$ question-answer pairs. We also used the Fish-500 dataset for the task of trait counting, resulting in a total of  $1K$ question-answer pairs. Across all tasks, our dataset comprises approximately  $469K$ question-answer pairs for  $30K$ biological images (see Table \ref{tab:dataset-stat-main}). Additional details on data distribution and key statistics are provided in the Supplementary.

\par \noindent \textbf{VLM Baselines:} We consider the following VLM baselines: GPT-4V(ision) \cite{openai2023gpt}\footnote{We use \textit{gpt-4-1106-vision-preview} model as GPT-4v in our experiments.}, LLaVA-v1.5 (7B/13B) \cite{liu2023visual}, COG-VLM \cite{wang2023cogvlm}, MiniGPT-4 (Vicuna 7B/13B) \cite{zhu2023minigpt}, BLIP-FLAN-T5-XL/XXL \cite{li2023blip}, and INSTRUCT-BLIP (Vicuna 7B/13B) \cite{dai2023instructblip}. We used the latest checkpoints for each model available to date. We used the same question prompt for all models to ensure consistent comparison of results for a variety of open and multiple-choice (MC) questions across the five scientific tasks of our dataset. All the experiments were conducted using NVIDIA A100 GPUs. See supplementary for more details of the VLM baselines.

\par \noindent \textbf{Evaluation Metrics:} We used micro-averaged accuracy as our evaluation metric for all experiments. We designed a systematic rule-based evaluation pipeline to evaluate VLM responses against the ground truths. For each question category, we provide the accuracy percentage of random choice as a basic baseline, where each possible answer is considered equally likely (yielding an accuracy of 25\% for MC questions with four choices).

\section{Results}

 


\begin{table*}[t]
\centering
\setlength\tabcolsep{2pt} 
\renewcommand{\arraystretch}{1.5}
\resizebox{\textwidth}{!}{
\begin{tabular}{ccrrrrrrrrrrrrr}
\toprule
\multicolumn{2}{c}{} & \multicolumn{13}{c}{\textbf{Models}} \\
\cmidrule(lr){3-15}
\textbf{Dataset }& \renewcommand{\arraystretch}{1.2}{\begin{tabular}[c]{@{}c@{}}\textbf{Question} \\ \textbf{type}\end{tabular}}  & \textit{gpt-4v} &  \renewcommand{\arraystretch}{1.2}{\begin{tabular}[c]{@{}c@{}}\textit{llava} \\ \textit{v1.5-7b}\end{tabular}} &  \renewcommand{\arraystretch}{1.2}{\begin{tabular}[c]{@{}c@{}}\textit{llava} \\ \textit{v1.5-13b}\end{tabular}} &  \renewcommand{\arraystretch}{1.2}{\begin{tabular}[c]{@{}c@{}}\textit{cogvlm} \\ \textit{chat}\end{tabular}} & \renewcommand{\arraystretch}{1.2}{\begin{tabular}[c]{@{}c@{}}\textit{BLIP} \\ \textit{flan-xl}\end{tabular}} &   \renewcommand{\arraystretch}{1.2}{\begin{tabular}[c]{@{}c@{}}\textit{BLIP} \\ \textit{flan-xxl}\end{tabular}} &  \renewcommand{\arraystretch}{1.2}{\begin{tabular}[c]{@{}c@{}}\textit{minigpt4} \\ \textit{vicuna-7B}\end{tabular}} &  \renewcommand{\arraystretch}{1.2}{\begin{tabular}[c]{@{}c@{}}\textit{minigpt4} \\ \textit{vicuna-13B}\end{tabular}} & \renewcommand{\arraystretch}{1.2}{\begin{tabular}[c]{@{}c@{}}\textit{instruct} \\ \textit{flant5xl}\end{tabular}} &  \renewcommand{\arraystretch}{1.2}{\begin{tabular}[c]{@{}c@{}}\textit{instruct} \\ \textit{flant5xxl}\end{tabular}} &  \renewcommand{\arraystretch}{1.2}{\begin{tabular}[c]{@{}c@{}}\textit{instruct} \\ \textit{vicuna7B}\end{tabular}} &  \renewcommand{\arraystretch}{1.2}{\begin{tabular}[c]{@{}c@{}}\textit{instruct} \\ \textit{vicuna13B}\end{tabular}} & \renewcommand{\arraystretch}{1.2}{\begin{tabular}[c]{@{}c@{}}\textit{Random} \\ \textit{Choice}\end{tabular}}  \\
\midrule
\multicolumn{15}{c}{\textbf{Species Classification}}\\
\midrule
\multirow{2}{*}{\textbf{Fish-10K}} & Open & 1.01 & \cellcolor{blue!35}2.32 & 0.40 & 0.11 & 0.01 & \cellcolor{blue!15}1.59 & 0.50 & 0.38 & \cellcolor{red!35}0.00 & 1.46 & \cellcolor{red!35}0.00 & \cellcolor{red!35}0.00 & 0.20\\
& MC & \cellcolor{blue!15}35.91 & \cellcolor{blue!35}40.20 & 32.27 & 31.72 & 29.76 & 33.36 & 29.02 & 27.45 & 30.86 & 31.70 & \cellcolor{red!15}27.27 & \cellcolor{red!35}26.57 & 25.00\\

\hline
\multirow{2}{*}{\textbf{Bird-10K}} & Open & \cellcolor{blue!35}17.40 & 1.45 & 2.06 & 0.86 & \cellcolor{red!35}0.00 & 0.57 & \cellcolor{blue!15}2.80 & 2.56 & \cellcolor{red!35}0.00 & 0.50 & 0.07 & \cellcolor{red!35}0.00 & 0.53 \\
& MC & \cellcolor{blue!35}82.58 & 50.32 & \cellcolor{blue!15}55.36 & 44.73 & 33.68 & 34.75 & \cellcolor{red!35}23.95 & \cellcolor{red!15}27.62 & 36.36 & 35.83 & 44.00 & 46.55 & 25.00\\

\hline
\multirow{2}{*}{\textbf{Butterfly-10K}} & Open & 0.04 & 0.05 & \cellcolor{red!35}0.00 & 0.01 & \cellcolor{red!35}0.00 & \cellcolor{red!35}0.00 & \cellcolor{blue!15}0.07 & 0.01 & \cellcolor{red!35}0.00 & \cellcolor{red!35}0.00 & \cellcolor{blue!35}9.94 & \cellcolor{red!35}0.00 & 1.54\\
& MC & 28.91 & \cellcolor{blue!35}50.24 & \cellcolor{blue!15}44.58 & 36.45 & \cellcolor{red!35}25.14 & 28.88 & 33.06 & 28.90 & \cellcolor{red!15}25.28 & 36.67 & 41.70 & 34.48 & 25.00	\\
\midrule
\multicolumn{15}{c}{\textbf{Trait Identification}}\\
\midrule
\textbf{Fish-10K} & MC & \cellcolor{blue!35}82.18 & 56.84 & 45.15 & 46.92 & 68.36 & 39.33 & 55.08 & 51.87 & 64.34 &
       \cellcolor{red!15}39.26 & \cellcolor{blue!15}81.95 & \cellcolor{red!35}20.69 & 50.0\\

\hline
\textbf{Bird-10K} & MC & 62.22 & \cellcolor{red!35}34.68 & 46.14 & 63.93 & 50.11 & 41.38 & \cellcolor{red!15}39.11 & 40.44 & 47.89 & 45.52 & \cellcolor{blue!15}77.91 & \cellcolor{blue!35}89.98 & 31.12 \\

\midrule
\multicolumn{15}{c}{\textbf{Trait Grounding}}\\
\midrule
\textbf{Fish-500} & MC & \cellcolor{blue!15}29.41 & 24.87 & \cellcolor{red!15}17.98 & 23.42 & 23.32 & 25.14 & 22.18 & 25.58 & \cellcolor{red!35}7.20 & 27.09 & \cellcolor{blue!35}33.51 & 26.90 & 25.00\\

\hline
\textbf{Bird-500} & MC & 8.1  & 26.92 & \cellcolor{blue!35}35.36 & 23.2  &  11.83   & 10.52 & 15.39 & 24.22 &  \cellcolor{red!15}3.48 &
        \cellcolor{red!35}0.81 & \cellcolor{blue!15}30.24 & 13.91 & 25.00 \\

\midrule
\multicolumn{15}{c}{\textbf{Trait Referring}}\\
\midrule
\textbf{Fish-500} & MC & 28.15 & 27.07 & 29.14 & 28.19 & \cellcolor{red!35}24.93 & \cellcolor{red!15}25.68 & \cellcolor{blue!35}39.24 & 31.21 & 31.75 & 25.78 & 28.04 & \cellcolor{blue!15}32.73 & 25.00\\

\hline
\textbf{Bird-500} & MC & \cellcolor{blue!35}42.28 & 30.5  & 29.64 & \cellcolor{red!35}18.45 & 35.16 & 40.59 & 26.04 & 35.88 & 27.52 & \cellcolor{blue!15}41.69 & 23.03 & \cellcolor{red!15}22.69  & 25.00 \\

\midrule
\multicolumn{15}{c}{\textbf{Trait Counting}}\\
\midrule
\multirow{2}{*}{\textbf{Fish-500}} & Open & 16.4  & 47.4  & 52.0   & 14.8  & 37.6  & \cellcolor{blue!35}63.4  & \cellcolor{red!15}13.6  & 31.53 & 50.2  & \cellcolor{blue!15}61.4  & \cellcolor{blue!15}61.4  &  \cellcolor{red!35}0.0 & 25.00 \\
 & MC & 44.80 &  \cellcolor{red!15}13.20 &  54.80 & 21.00   &  64.8  &  \cellcolor{blue!35}78.2  &  22.00 &   25.00   & \cellcolor{blue!15}74.0   &  69.4  & 15.80   & \cellcolor{red!35}11.80 & 25.00 \\
\midrule
\multicolumn{2}{c}{\textit{Overall}} & \cellcolor{blue!35}34.24 & 29.0  & 31.78   & 25.27  & 28.91  & 30.24  & \cellcolor{red!15}23.0  & 25.19 & 28.49  & 29.79  & \cellcolor{blue!15}33.92  &  \cellcolor{red!35}23.31 & 22.03\\

\bottomrule
\end{tabular}
}
\caption{Zero-shot accuracy comparison of VLM baselines (in \% ranging from 0 to 100) for the five scientific tasks. 
Results are color-coded as \colorbox{blue!35} {Best}, \colorbox{blue!15} {Second best}, \colorbox{red!35} {Worst}, \colorbox{red!15} {Second worst}.
}
\label{tab:zero-shot-eval}

\end{table*}

 Table \ref{tab:zero-shot-eval} compares the accuracies of VLMs in percentages (ranging from 0 to 100) across the five tasks and over multiple organism datasets.
 We make the following observations from this result.

\par \noindent \textbf{All VLMs show poor accuracy on open questions but perform better on MC questions.} The zero-shot species classification accuracy of all VLMs on open-ended questions is notably weaker than MC questions. Even the best-performing models, LLaVA-13B, GPT-4V, and Instruct-Vicuna-7B, only achieve accuracies of 2.32\%, 17.46\%, and 3.62\%, respectively, across the three organism datasets. This indicates a significant gap in the ability of existing VLMs to capture the scientific knowledge necessary to differentiate between species (often requiring subtle or nuanced features) without being provided with candidate answer choices. Open-ended species classification is particularly hard for pre-trained VLMs that are not typically trained to provide scientific names of organisms (e.g., \textit{Lepomis cyanellus}) rather than providing their common names (e.g., \textit{green sunfish}). However,  the inclusion of candidate answers (or options) in the question prompt serves as a helpful clue to VLMs for narrowing down the solution space and finding the correct answer potentially using elimination strategies. While VLMs are able to utilize these additional hints and work their way through to the correct answer in MC questions, note that open questions are practically more relevant to scientists operating in real-world settings.



\par \noindent \textbf{Bird dataset shows better accuracy than Fish or Butterfly datasets.} Most VLMs show significantly better performance on the Bird-$10K$ dataset in comparison to the Fish-$10K$ and Butterfly-$10K$ datasets. For example, the highest accuracy across all VLMs on the Bird-$10K$ dataset is 82.58\%, while it is 40.20\% and 50.24\% on the Fish-$10K$ and Butterfly-$10K$ datasets, respectively. A potential reason is that while the bird dataset is a subset of the CUB dataset \cite{wah2011cub} that is commonly used in machine learning literature and has images with natural in-the-wild backgrounds, the butterfly and fish datasets contain images of specimens preserved in museum collections with artificial backgrounds and with imaging artifacts that are not typical for large-scale computer vision datasets. We hypothesize that many of the pre-trained VLM baselines may have seen images similar to those in the Bird dataset during training, leading to their better performance.




\par \noindent \textbf{Can VLMs effectively identify biological traits?} The performance of most VLMs in trait identification appears significantly better than their performance in species classification, with GPT-4V reaching 82.18\% accuracy on the Fish-$10K$ dataset and Instruct-Vicuna-13B achieving 89.98\% on Bird-$10K$. However, some traits such as ``eye'', ``head'', and ``mouth'' are almost always present in every organism image, so simply answering ``yes, the trait is present'' can lead to high accuracy in trait identification.
In contrast to the fish dataset, the bird dataset poses more intricate questions regarding a variety of multi-class traits that require a nuanced understanding of colors, patterns, and physical trait dimensions, such as the color of the bill, wing patterns, and tail shapes. 


\begin{figure*}[t]
    \centering
    \includegraphics[width=\linewidth]{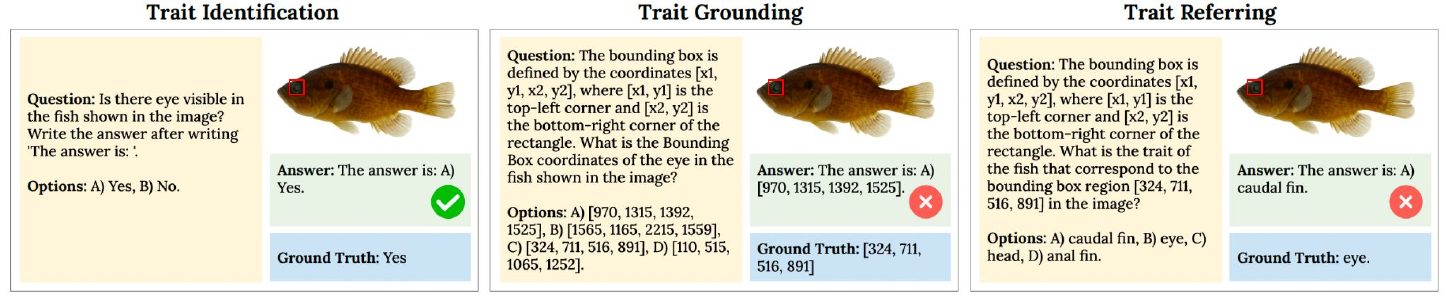}
    \caption{Examples of correct and incorrect predictions of GPT-4V for trait identification, trait grounding, and trait-referring tasks related to the ``eye''. For visualization assistance, a \textcolor{red}{red-colored bounding box} is added around the ``eye'' in the image.}
    \label{fig:identification-vs-detection}
\end{figure*}


\par \noindent \textbf{VLMs struggle 
in localizing traits in images.} While most VLMs perform well on the task of Trait Identification, it is crucial to determine if they are focusing on the correct image regions to answer trait-related questions. We thus analyze the performance of VLMs on the tasks of trait grounding (i.e., \textit{text to location}) and trait referring (i.e., \textit{location to text}).  We can see that there is a significant drop in the accuracy of trait grounding and referring tasks compared to the trait identification task. This shows that while VLMs can potentially leverage knowledge of trait choices to identify traits, they struggle in localizing the traits in the image and thus visually ground their reasoning. Figure \ref{fig:identification-vs-detection} shows an illustrative example of GPT-4V prediction where it predicts the presence of the trait ``eye'' correctly but fails to localize it in grounding and referring tasks.


\par \noindent \textbf{Counting biological traits is difficult for VLMs.}
Recent studies \cite{qin2023good, jiang2024effectiveness, yang2023dawn} have explored the gap in the ability of VLMs to count objects, which is aligned with our results in Table \ref{tab:zero-shot-eval}. All VLMs, except for BLIP-flan-T5-XXL, show lower performance in counting traits, despite performing well on the trait identification task. The overall average accuracy for the VLMs is displayed in the last block, with GPT-4V(ision) exhibiting the best performance.

We further analyze the errors of different VLMs to better understand their behavior. We find that GPT-4V shows a reduced rate of incorrect responses but a higher incidence of ``Other'' responses, which include apologetic expressions, admissions of inability to precisely visualize the organism, and disclaimers regarding lack of expert guidance (see Supplementary for more details).

\subsection{Analyzing the Role of Answer Choices in MC Questions on VLM Performance}
\label{sec:answer_choice_role}


Table \ref{tab:zero-shot-eval} showed that VLMs perform drastically better on MC questions compared to Open questions for species classification. A potential hypothesis for this observation is that VLMs are able to avoid incorrect answer choices (or options) that are too different from the correct option and thus are easy to eliminate. To test this hypothesis, 
we create three variants of the MC questions for species classification---easy, medium, and hard----where species choices in each variant have varying degrees of similarity determined by their taxonomic groupings.
In particular, note that the scientific name of an organism contains taxonomic information at three levels: \texttt{<genus name> <species name> <subspecies name>}\footnote{We only have subspecies level information for the Butterfly-$10K$ dataset.}. 
Since organisms that share taxonomic information have similar appearances, it is hard to differentiate species choices if they are from the same taxonomic group. On the other hand, it is easier to work with species choices from different taxonomic groups.
Hence, for the easy set, we selected 50 species from different genera, ensuring that all species choices appear quite different from each other. For the medium set, we increased the complexity by constructing species choices from the same genus but from 10 different species. The hard set presented the highest difficulty level for the butterfly dataset, with the answer choices being from the same genus and species but from 10 subspecies. Each difficulty level consists of 200 images from each set of organisms. 

\begin{table*}[t]
\centering
\setlength\tabcolsep{0.06cm} 
\renewcommand{\arraystretch}{1.5}
\resizebox{\textwidth}{!}{
\begin{tabular}{ccccccccccccccccc}
\toprule
\multicolumn{2}{c}{} & \multicolumn{15}{c}{\textbf{Models}} \\
\cmidrule(lr){3-17}
\textbf{Dataset }& \textbf{Difficulty }  & \textit{gpt-4v} & \textit{gpt-4o} & \renewcommand{\arraystretch}{1.2}{\begin{tabular}[c]{@{}c@{}}\textit{llava} \\ \textit{v1.5-7b}\end{tabular}} &  \renewcommand{\arraystretch}{1.2}{\begin{tabular}[c]{@{}c@{}}\textit{llava} \\ \textit{v1.5-13b}\end{tabular}} &  \renewcommand{\arraystretch}{1.2}{\begin{tabular}[c]{@{}c@{}}\textit{cogvlm} \\ \textit{chat}\end{tabular}} & \renewcommand{\arraystretch}{1.2}{\begin{tabular}[c]{@{}c@{}}\textit{BLIP} \\ \textit{flan-xl}\end{tabular}} &   \renewcommand{\arraystretch}{1.2}{\begin{tabular}[c]{@{}c@{}}\textit{BLIP} \\ \textit{flan-xxl}\end{tabular}} &  \renewcommand{\arraystretch}{1.2}{\begin{tabular}[c]{@{}c@{}}\textit{minigpt4} \\ \textit{vicuna-7B}\end{tabular}} &  \renewcommand{\arraystretch}{1.2}{\begin{tabular}[c]{@{}c@{}}\textit{minigpt4} \\ \textit{vicuna-13B}\end{tabular}} & \renewcommand{\arraystretch}{1.2}{\begin{tabular}[c]{@{}c@{}}\textit{instruct} \\ \textit{flant5xl}\end{tabular}} &  \renewcommand{\arraystretch}{1.2}{\begin{tabular}[c]{@{}c@{}}\textit{instruct} \\ \textit{flant5xxl}\end{tabular}} &  \renewcommand{\arraystretch}{1.2}{\begin{tabular}[c]{@{}c@{}}\textit{instruct} \\ \textit{vicuna7B}\end{tabular}} &  \renewcommand{\arraystretch}{1.2}{\begin{tabular}[c]{@{}c@{}}\textit{instruct} \\ \textit{vicuna13B}\end{tabular}} & \textit{CLIP}& \textit{BioCLIP} \\

\midrule
\multirow{2}{*}{\textbf{Fish}} & Easy & 44.50 & 37.50 & \colorbox{blue!15}{47.50} & 46.00 & \colorbox{red!35}{24.00} & 34.00 & \colorbox{red!15}{27.50} & 29.00 & 19.50 & 32.00 & 28.00 & 33.50 & 33.50 & 36.50 & \colorbox{blue!35}{55.50}\\
 & Medium & \colorbox{red!35}{3.50} & \colorbox{red!35}{5.50} & \colorbox{blue!35}{30.00} & 28.50 & 27.00 & 26.00 & 23.00 & 26.50 & 25.00 & 28.50 & 24.50 & 26.00 & 25.50 & 26.00 & \colorbox{blue!15}{29.00}\\

\midrule
\multirow{2}{*}{\textbf{Bird}} & Easy & \colorbox{blue!15}{73.50} & 68.00 & 53.50 & 50.00 & 38.50 & 34.50 & 36.00 & \colorbox{red!35}{21.00} & \colorbox{red!15}{32.00} & 41.00 & 33.00 & 43.50 & 39.00 & 57.00 & \colorbox{blue!35}{94.00}\\
 & Medium & \colorbox{blue!15}{41.00} & 40.50 & 30.50 & 37.00 & 30.00 & 25.50 & \colorbox{red!35}{21.00} & \colorbox{red!35}{21.00} & 24.00 & 27.00 & 27.00 & 24.50 & 26.50& 31.00 & \colorbox{blue!35}{95.00}\\
\midrule
\multirow{3}{*}{\textbf{Butterfly}} & Easy & \colorbox{red!15}{18.50} & \colorbox{red!35}{17.50} & 19.00 & 20.50 & 24.50 & 30.00 & 25.00 & \colorbox{blue!15}{34.50} & 26.00 & 24.50 & 22.50 & 19.00 & 24.50 & 21.50 & \colorbox{blue!35}{65.50}\\
 & Medium & \colorbox{red!35}{5.50} & \colorbox{red!15}{7.00} & 29.50 & 29.00 & 29.50 & 20.00 & 25.50 & \colorbox{blue!15}{33.00} & 25.00 & 27.50 & 25.00 & 25.00 & 25.00 & 21.50 & \colorbox{blue!35}{58.00}\\
 & Hard & \colorbox{red!15}{2.00} & \colorbox{red!35}{1.50} & 22.00 & 21.00 & \colorbox{blue!15}{32.00} & 26.50 & 20.00 & 29.50 & 24.00 & 22.50 & 24.00 & 24.00 & 21.00 & 21.50 & \colorbox{blue!35}{35.00}\\

\bottomrule
\end{tabular}}
\caption{Zero-Shot accuracy comparison for \textit{easy, medium, and hard} datasets. Results are color-coded as \colorbox{blue!35} {Best}, \colorbox{blue!15} {Second best}, \colorbox{red!35} {Worst}, \colorbox{red!15} {Second worst}.}
\label{tab:easy-medium-hard}

\end{table*}

Table \ref{tab:easy-medium-hard} shows the accuracies of the baseline VLMs for the easy, medium, and hard organism datasets. The pretrained VLMs generally perform best on the easy set and worst on the hard set for each organism. Moreover, there is a gradual improvement in the VLM performance from hard to easy questions. This suggests that the difficulty level of candidate answers (or options) in the question prompt significantly impacts VLMs' performance. Additionally, this outcome indicates that even SOTA VLMs have limitations in handling fine-grained queries. Table \ref{tab:easy-medium-hard} shows that GPT-4V and OpenAI's recent release GPT-4o do not perform well when tested on the medium and hard datasets for Fish and Butterfly. Due to this, we further analyze the errors of different VLMs to better understand their behavior. We provide the report in the Supplementary.

\subsection{Comparing Pre-trained VLMs with a Biologically Fine-tuned Model} 
We compare BioCLIP \cite{stevens2023bioclip}, a state-of-the-art foundation model for species classification fine-tuned with biological images and taxonomic names (TreeOfLife-10M dataset), with the pretrained VLMs. We observe that BioCLIP significantly outperforms large pretrained VLMs on the Bird-$10K$ and Butterfly datasets, suggesting that BioCLIP has been trained on images that are similar to the organisms present in these datasets. By comparing BioCLIP with CLIP, we can also see that fine-tuning foundation models with biological data provides large gains in classification performance. 
This suggests that the performance of SOTA VLMs can be further improved by fine-tuning on VLM4Bio Dataset. 
Further details comparing BioCLIP with SOTA VLMs are provided in the Supplementary.

\subsection{Analyzing Effects of Prompting on VLM Performance}

We considered three prompting techniques: Contextual Prompting, Dense Caption Prompting, and zero-shot Chain of Thought Prompting. For \textbf{Contextual prompting}, we provided a single-line description (context) of the tasks (e.g., we add ``\textit{Each biological species has a unique scientific name composed of two parts: the first for the genus and the second for the species within that genus.}'' before the species classification question to give some additional context on the task). \textbf{ Dense Caption prompting} involves two stages: (1) first, we prompt the VLM to generate a dense caption of the specimen image such that the caption contains all the necessary trait information of the specimen. (2) We add the dense caption before the question and prompt ``\textit{Use the above dense caption and the image to answer the following question.}'' to generate responses from the VLM. Similarly, the \textbf{Zero-Shot Chain-of-Thought (CoT)} happens in two stages: (1) First, we prompt the VLM to generate the reasoning for a given VQA and multiple choices (options). Zero-shot CoT appends ``\textit{Let's think step by step.}'' after the question and options to generate the reasoning. (2) We then add the reasoning after the VQA and prompt ``\textit{Please consider the following reasoning to formulate your answer}'' to generate the VLM response. 
We curated a prompting dataset of $500$ multiple-choice (MC) VQAs for each set of organisms, which is a subset of the VLM4Bio dataset for species classification.  

\begin{table*}[t]
\centering
\setlength\tabcolsep{0.06cm} 
\renewcommand{\arraystretch}{0.9}
\resizebox{0.7\textwidth}{!}{
\begin{tabular}{ccrrrrrrr}
\toprule
\multicolumn{2}{c}{} & \multicolumn{7}{c}{\textbf{Models}} \\
\cmidrule(lr){3-9}
\textbf{Dataset }& \textbf{Prompting}  & \textit{gpt-4v} & \textit{gpt-4o} &  \renewcommand{\arraystretch}{1.2}{\begin{tabular}[c]{@{}c@{}}\textit{llava} \\ \textit{v1.5-7b}\end{tabular}} &  \renewcommand{\arraystretch}{1.2}{\begin{tabular}[c]{@{}c@{}}\textit{llava} \\ \textit{v1.5-13b}\end{tabular}} &  \renewcommand{\arraystretch}{1.2}{\begin{tabular}[c]{@{}c@{}}\textit{cogvlm} \\ \textit{chat}\end{tabular}} & \renewcommand{\arraystretch}{1.2}{\begin{tabular}[c]{@{}c@{}}\textit{BLIP} \\ \textit{flan-xl}\end{tabular}} &   \renewcommand{\arraystretch}{1.2}{\begin{tabular}[c]{@{}c@{}}\textit{BLIP} \\ \textit{flan-xxl}\end{tabular}} \\

\midrule
\multirow{4}{*}{\textbf{Fish-Prompting}} & No Prompting & 34.40 & 79.00 & 41.60 & 35.40 & 31.00 & 28.60 & 22.60 \\
 & Contextual & 30.00 & 77.20 & 40.20 & 35.60 & 25.60 & 27.20 & 26.60 \\
 & Dense Caption & \colorbox{red!15}{18.80} & 78.60 & 26.00 & 27.60 & 32.00 & 28.40 & 29.80 \\
 & CoT & 42.60 & \colorbox{blue!15}{86.00} & 41.40 & 34.80 & 26.80 & 29.20 & 24.60\\
\midrule
\multirow{4}{*}{\textbf{Bird-Prompting}} & No Prompting & 78.80 & 97.60 & 44.20 & 49.80 & 45.40 & 35.60 & 35.80\\
 & Contextual & 78.60 & \colorbox{blue!15}{98.60} & 44.00 & 52.00 & 49.40 & 35.60 & 30.40\\
 & Dense Caption & 87.40 & 97.00 & 33.40 & 41.00 & 44.00 & 25.60 & \colorbox{red!15} {22.80}\\
 & CoT & 62.60 & \colorbox{blue!15}{98.60} & 37.40 & 47.80 & 42.20 & 30.60 & 31.00\\
\midrule

\multirow{4}{*}{\textbf{Butterfly-Prompting}} & No Prompting & 13.20 & 56.40 & 27.20 & 26.80 & 25.60 & 24.40 & 21.20 \\
 & Contextual & \colorbox{red!15}{9.20} & 56.20 & 26.00 & 24.60 & 27.20 & 23.60 & 24.60 \\
 & Dense Caption & 49.60 & 63.20 & 25.20 & 23.80 & 27.00 & 23.20 & 23.20\\
 & CoT & 63.60 & \colorbox{blue!15}{74.60} & 21.40 & 23.20 & 34.60 & 37.20 & 23.60\\

\bottomrule
\end{tabular}}
\caption{ Zero-shot accuracy comparison for different prompting techniques of seven VLMs (in \% ranging from 0 to 100). Results are color-coded as \colorbox{blue!15} {Best} and \colorbox{red!15} {Worst}.
}
\label{tab:prompting}
\end{table*}



Table \ref{tab:prompting} compares best-performing VLMs on the prompting dataset. The CoT rows of the table demonstrate that only GPT-4V and GPT-4o have reasoning capabilities that can significantly improve their response to biological questions, while smaller models like LLaVa and BLIP do not show much improvement. Furthermore, providing extra context and caption is more useful for GPT-4V and GPT-4o than the smaller models. This resonates with the findings from \cite{zhang2023multimodal} that the reasoning abilities of VLMs only emerge after a certain model size. The success of Dense Caption prompting and CoT prompting depends on how well they generate the dense caption or the reasoning in the first stage. 
We report example prompts with VLM responses as case studies in the Supplementary.


\begin{table*}[t]
\centering
\setlength\tabcolsep{0.2cm} 
\renewcommand{\arraystretch}{0.9}
\resizebox{0.9\textwidth}{!}{
\begin{tabular}{ccrrrrrrr}
\toprule
\multicolumn{2}{c}{} & \multicolumn{7}{c}{\textbf{Models}} \\
\cmidrule(lr){3-9}
\textbf{Dataset }& \textbf{Metrics}  & \textit{gpt-4v} & \textit{gpt-4o} &  \begin{tabular}[c]{@{}c@{}}\textit{llava} \\ \textit{v1.5-7b}\end{tabular} &  \begin{tabular}[c]{@{}c@{}}\textit{llava} \\ \textit{v1.5-13b}\end{tabular} &  \begin{tabular}[c]{@{}c@{}}\textit{cogvlm} \\ \textit{chat}\end{tabular} & \begin{tabular}[c]{@{}c@{}}\textit{BLIP} \\ \textit{flan-xl}\end{tabular} &   \begin{tabular}[c]{@{}c@{}}\textit{BLIP} \\ \textit{flan-xxl}\end{tabular} \\

\midrule
\multicolumn{9}{c}{\textbf{False Confidence Test (FCT)}}\\
\midrule
\multirow{2}{*}{\textbf{Fish-Prompting}} & Accuracy & 34.20 & \colorbox{blue!15}{73.60} & 25.00 & 28.60 & 24.60 & \colorbox{red!15}{0.00} & 7.00 \\
 & Agreement Score & 4.40 & 16.60 & 99.80 & 19.20 & 74.40 & 0.00 & 28.4 \\
\midrule
\multirow{2}{*}{\textbf{Bird-Prompting}} & Accuracy & 73.40 & \colorbox{blue!15}{99.00} & 25.40 & 35.80 & 19.80 & \colorbox{red!15}{0.00} & 20.20\\
 & Agreement Score & 11.40 & 21.00 & 93.20 & 17.80 & 47.80 & 0.00 & 79.80\\
\midrule
\multirow{2}{*}{\textbf{Butterfly-Prompting}} & Accuracy & 5.20 & \colorbox{blue!15}{53.40} & 27.20 & 26.60 & 6.20 & \colorbox{red!15}{0.00} & 5.00 \\
 & Agreement Score & 2.60 & 12.40 & 95.40 & 5.60 & 13.80 & 0.00 & 19.00 \\
\midrule
\multicolumn{9}{c}{\textbf{None of the Above (NOTA) Test}}\\
\midrule
\multirow{1}{*}{\textbf{Fish-Prompting}} & Accuracy & \colorbox{blue!15}{81.40} & 44.80 & 3.40 & 3.80 & \colorbox{red!15}{0.00} & 4.00 & \colorbox{red!15}{0.00} \\
\midrule
\multirow{1}{*}{\textbf{Bird-Prompting}} & Accuracy & 75.00 & \colorbox{blue!15}{91.40} & 1.00 & 1.20 & \colorbox{red!15}{0.00} & 31.40 & \colorbox{red!15}{0.00}\\
\midrule
\multirow{1}{*}{\textbf{Butterfly-Prompting}} & Accuracy & 50.40 & 4.60 & 1.00 & 4.60 & \colorbox{red!15}{0.00} & \colorbox{blue!15}{51.00} & \colorbox{red!15}{0.00} \\
\bottomrule
\end{tabular}}
\vspace{1.5ex}
\caption{Performance of seven VLMs on the NOTA and FCT reasoning tests. Results are color-coded as \colorbox{blue!15} {Best} and \colorbox{red!15} {Worst}.
}
\label{tab:reasoning}
\end{table*}


\subsection{Analyzing Tests for Reasoning Hallucination}
To further understand whether pretrained VLMs can respond with logically coherent and factually accurate reasoning, we evaluate VLMs on two sets of reasoning for hallucination tests - \textbf{False Confidence Test (FCT)} and \textbf{None of the Above (NOTA) Test} - inspired by \cite{umapathi2023med}. 
For the FCT, we randomly select an option from the list of given choices and prompt it to the VLM as a ``suggested correct answer'' along with the question and options. To evaluate VLMs on FCT, we use Accuracy as well as the Agreement score, which is the percentage of times the VLM agrees with the suggested answer, irrespective of whether that is right or wrong. A high agreement score with a low overall accuracy indicates poor performance as it suggests that the model is simply following the suggestion either because of a lack of knowledge or low confidence in its own response. On the other hand, in the NOTA Test, we replace the correct option with \say{None of the Above}, requiring the model to produce \say{None of the above} for all the questions. 
From Table \ref{tab:reasoning}, we can see that LLaVa-v1.5-7B shows poor accuracy on both tests and high agreement score on FCT. Out of all the VLMs, GPT-4V and GPT-4o demonstrate the highest accuracy, i.e., the lowest false confidence. More details on the prompts and examples of the responses have been provided in the Supplementary.

\vspace{-1ex}
\section{Limitations}
Our work has three main limitations. First, while no prior VQA benchmark dataset exists for organismal biology to the best of our knowledge, we focused on only three organisms—fish, bird, and butterfly—out of the many available due to resource constraints. Adding more organisms with manually annotated trait data will require additional resources and domain expertise, which could be pursued in future work. Second, since it is not feasible to manually inspect all images to ensure that they are free from label noise, we acknowledge that some noise may be present in the labels used for evaluating models on our current dataset, which we plan to address in future iterations. Third, due to resource constraints, certain proprietary VLMs that require purchasing APIs like Gemini-Pro \cite{team2023gemini}, Gemini-Ultra \cite{team2023gemini}, and Claude Opus \cite{anthropic2024claude} were also not included in the evaluation. We anticipate that their performance will be comparable to that of the proprietary GPT-4V \cite{gpt4v} and GPT-4o \cite{gpt4o} considered in our evaluation.

\section{Conclusion and Future Work}
We presented VLM4Bio, a benchmark dataset to evaluate the zero-shot performance of pretrained VLMs on biologically relevant questions involving biodiversity images, exposing gaps in SOTA VLMs when applied to organismal biology. 
We observe that while VLMs are able to perform reasonably well on simpler tasks, e.g., using questions with multiple-choice formats and images with natural-looking backgrounds, they struggle in complex task settings that are practically more relevant to biologists. Through our study on prompting and reasoning tests on the VLM4Bio dataset, we observe that very large SOTA VLMs such as GPT-4V and GPT-4o have reasoning capabilities that can significantly improve the response to biological questions. We did not explore Retrieval Augmented Generation (RAG) \cite{lewis2020retrieval} or knowledge-infused prompting \cite{xu2023knowledge} techniques since they require additional knowledge bases, which could be developed in future work. Future works can also focus on finetuning VLMs on the VLM4Bio dataset instead of comparing zero-shot performance.

\section*{Acknowledgements}
This research is supported by National Science Foundation (NSF) awards for the HDR Imageomics Institute (OAC-2118240). We are thankful for the support of computational resources provided by the Advanced Research Computing (ARC) Center at Virginia Tech. This manuscript has been authored by UT-Battelle, LLC, under contract DE-AC05-00OR22725 with the US Department of Energy (DOE). The US government retains, and the publisher, by accepting the article for publication, acknowledges that the US government retains a nonexclusive, paid-up, irrevocable, worldwide license to publish or reproduce the published form of this manuscript or allow others to do so for US government purposes. DOE will provide public access to these results of federally sponsored research in accordance with the DOE Public Access Plan (\url{https://www.energy.gov/doe-public-access-plan}).

\bibliography{sample-base}
\bibliographystyle{unsrt}

\newpage
\appendix

\addtocontents{toc}{\protect\setcounter{tocdepth}{10}} 
\clearpage  
\begin{center}
\section*{\textcolor{blue}{VLM4Bio: Supplementary Material}}
\end{center}

\renewcommand{\contentsname}{Table of Contents}
\tableofcontents

\begin{figure}[h!]
    \centering
    \begin{subfigure}[b]{0.40\textwidth}
        \centering
        \includegraphics[width=\textwidth]{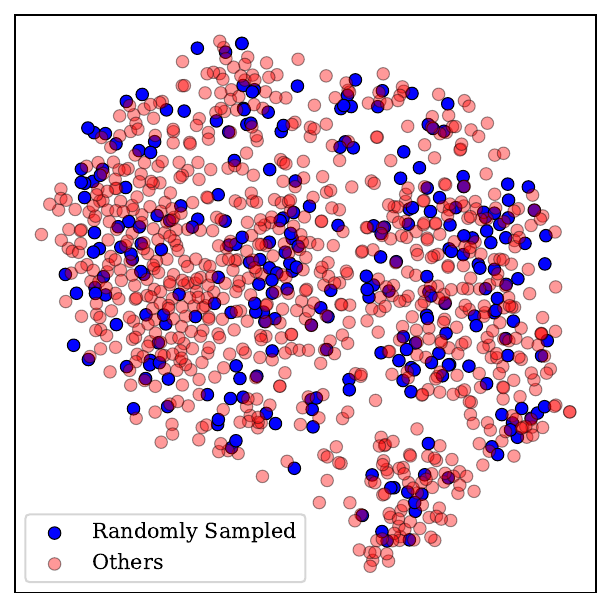}
        \caption{\textit{Lepomis humilis}}
        \label{fig:fish_tsne_1}
    \end{subfigure}
    \begin{subfigure}[b]{0.40\textwidth}
        \centering
        \includegraphics[width=\textwidth]{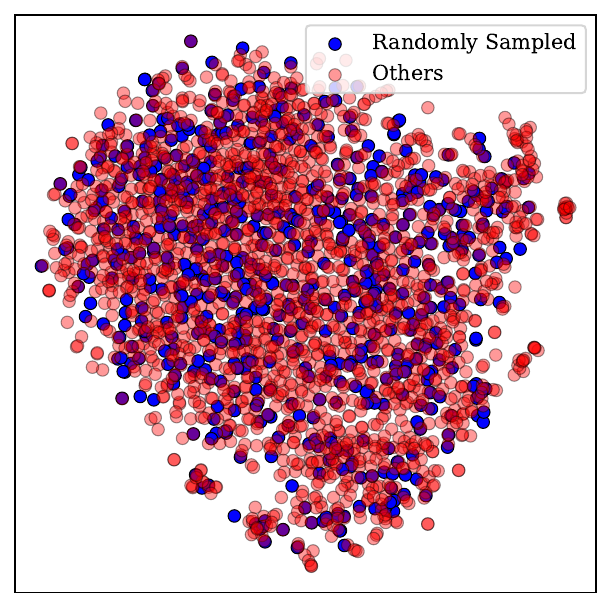}
        \caption{\textit{Lepomis cyanellus}}
        \label{fig:fish_tsne_2}
    \end{subfigure}
    \begin{subfigure}[b]{0.40\textwidth}
        \centering
        \includegraphics[width=\textwidth]{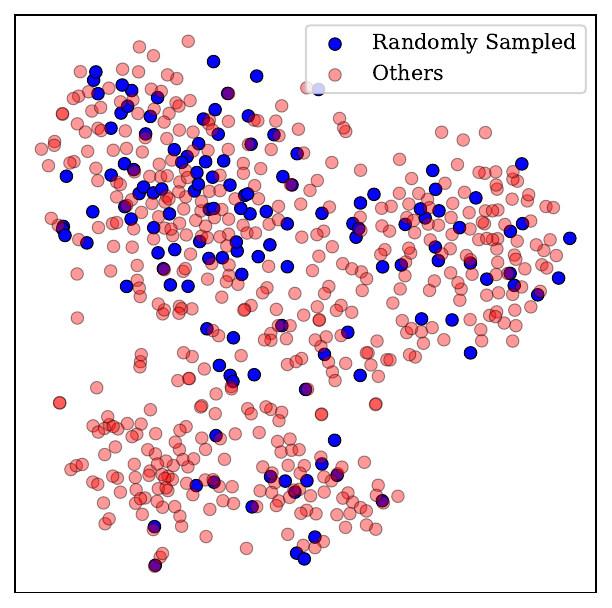}
        \caption{\textit{Noturus gyrinus}}
        \label{fig:fish_tsne_3}
    \end{subfigure}
    \begin{subfigure}[b]{0.40\textwidth}
        \centering
        \includegraphics[width=\textwidth]{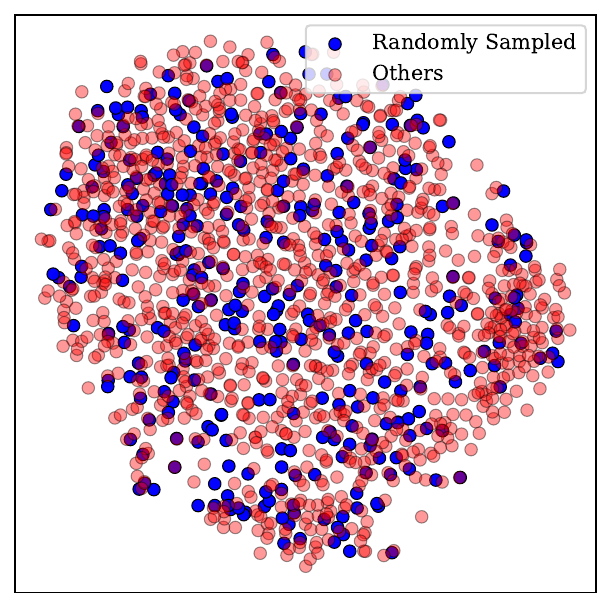}
        \caption{\textit{Lepomis megalotis}}
        \label{fig:fish_tsne_4}
    \end{subfigure}
    \begin{subfigure}[b]{0.40\textwidth}
        \centering
        \includegraphics[width=\textwidth]{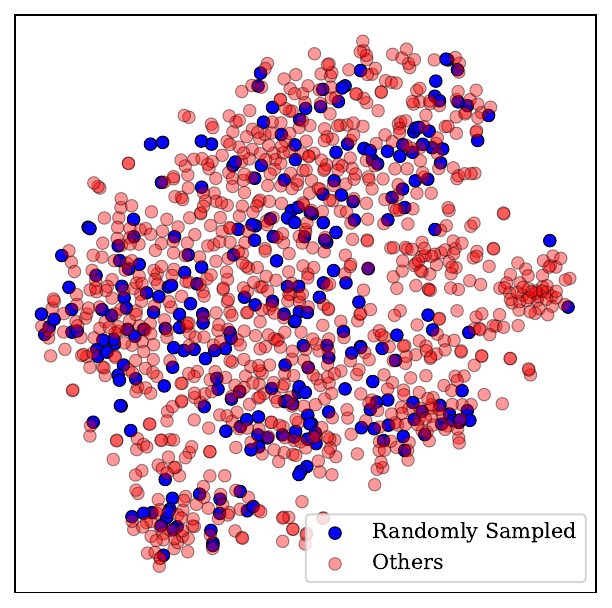}
        \caption{\textit{Phenacobius mirabilis}}
        \label{fig:fish_tsne_5}
    \end{subfigure}
    \begin{subfigure}[b]{0.40\textwidth}
        \centering
        \includegraphics[width=\textwidth]{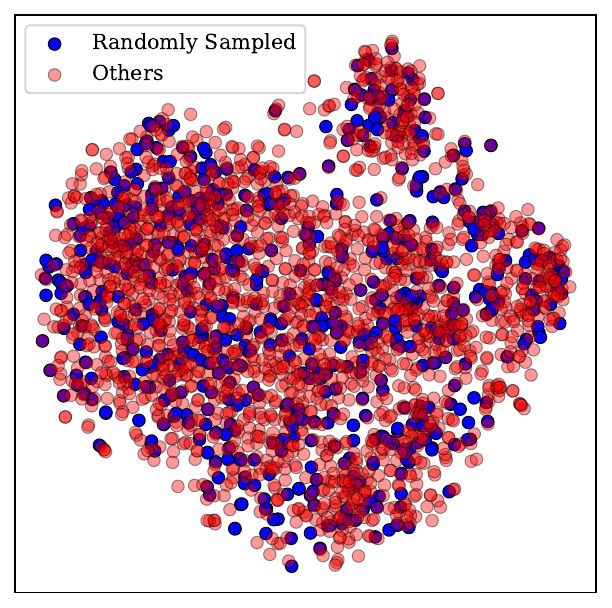}
        \caption{\textit{Lepomis macrochirus}}
        \label{fig:fish_tsne_6}
    \end{subfigure}
    \caption{t-SNE plots to illustrate the effectiveness of random sampling with the majority species in the Fish-10K dataset. Randomly sampled images are shown as blue dots, while the remaining data points are represented by red dots. Subcaptions display the scientific names of the corresponding species. To generate the vector representation of the images, we leverage a VGG19 pretrained on the ImageNet dataset.}
    \label{fig:tsne}
\end{figure}

\section{Dataset Preprocessing}
We collected images of three taxonomic groups of organisms: fish, birds, and butterflies, each containing around $10K$ images. Images for fish (\textbf{Fish-10K}) were curated from the larger image collection, FishAIR \cite{fishair}, which contains images from the Great Lakes Invasive Network Project (GLIN) \cite{GLIN} and Integrated Digitized Biocollections (iDigBio) \cite{iDigBio}. These images originate from various museum collections such as INHS \cite{inhs}, FMNH \cite{fmnh}, OSUM \cite{osum}, JFBM \cite{JFBM}, UMMZ \cite{UMMZ} and UWZM \cite{uwzm}. We created the Fish-10K dataset by randomly sampling $10K$ images and preprocessing the images to crop and remove the background. 

To ensure diversity within the Fish-10K dataset, we applied a targeted sampling strategy in the source collection, FishAIR \cite{fishair}. Specifically, we retained all images of species with fewer than 200 images, considering these as minority or rare classes. Random sampling was applied only to the majority species—those with more than 200 images per class. To assess the potential sampling bias among the majority species, we generated feature vectors for each image in Fish-10K using a pretrained VGG-19 model. In Figure \ref{fig:tsne}, we present species-wise t-SNE plots of these feature vectors for several majority species. Our analysis shows that the distribution of sampled images closely mirrors the distribution of images that were not included in the dataset (denoted as ``others'' in the plot). This suggests that our random sampling approach provides a sufficiently accurate representation of the original distribution for the majority species. For consistency, we leverage GroundingDINO \cite{liu2023grounding} to crop the fish body from the background and Segment Anything Model (SAM) \cite{kirillov2023segment} to remove the background. The Fish-10K dataset contains images of specimens preserved in museum collections with artificial backgrounds with imaging artifacts that are not typical for large-scale computer vision datasets. Moreover, these backgrounds can introduce unexpected bias. Hence, we removed the backgrounds using SAM to create a controlled environment for our experiments.

We curated the images for butterflies (\textbf{Butterfly-10K}) from the Jiggins Heliconius Collection dataset \cite{lawrence_campolongo_j2024}, which has images collected from various sources \footnote{Sources: \cite{but1_gabriela_montejo_kovacevich_2020_4289223, but2_patricio_a_salazar_2020_4288311, but3_montejo_kovacevich_2019_2677821, but4_jiggins_2019_2682458, but5_montejo_kovacevich_2019_2682669, but6_montejo_kovacevich_2019_2684906, but7_warren_2019_2552371, but8_warren_2019_2553977, but9_montejo_kovacevich_2019_2686762, but10_jiggins_2019_2549524, but11_jiggins_2019_2550097, but12_joana_i_meier_2020_4153502, but13_montejo_kovacevich_2019_3082688, but14_montejo_kovacevich_2019_2813153, but15_salazar_2018_1748277, but16_montejo_kovacevich_2019_2702457, but17_salazar_2019_2548678, but18_pinheiro_de_castro_2022_5561246, but19_montejo_kovacevich_2019_2707828, but20_montejo_kovacevich_2019_2714333, but21_gabriela_montejo_kovacevich_2020_4291095, but22_montejo_kovacevich_2021_5526257, but23_warren_2019_2553501, but24_salazar_2019_2735056, but25_mattila_2019_2555086}}. We carefully sampled $10K$ images for Butterfly-10K from the entire collection to ensure the images capture unique specimens and represent a diverse set of species by adopting the following two steps.
\textbf{First}, the butterfly images show various angles, including dorsal and ventral views, forewing dorsal and ventral views, and hindwing dorsal and ventral views. To ensure consistency, we only selected images with dorsal view and removed all images of hybrid species. \textbf{Second}, we further filtered the dataset based on the unique specimen ID to ensure no specimen was repeated more than once. For species with more than 2000 images, we performed random sampling (no sampling was performed for species with sizes less than 2000). We ensure each species has
a minimum of 20 images and no more than 2,000 images. The Butterfly-10K dataset contains a significant number of images of \textit{Heliconius melpomene} and \textit{Heliconius erato} species. We utilized the subspecies information of these two species to create a hard dataset for analyzing the impact of answer choices on VLM performance, as described in Section \ref{sec:answer_choice_role}.

The images for birds (\textbf{Bird-10K}) are obtained from the CUB-200-2011 \cite{wah2011caltech} dataset by taking 190 species for which the common name to scientific name mapping is available. This results in a fairly balanced dataset with around $11K$ images in total.

The scientific names for the images of Fish-10K and Butterfly-10K were obtained directly from their respective sources. For Bird-10K, we obtained the scientific names from the iNatLoc500 \cite{iNatLoc500} dataset. We curated around  $31K$ question-answer pairs in both open and multiple-choice (MC) question formats for evaluating species classification tasks. The species-level trait presence/absence matrix for Fish-10K was manually curated with the help of biological experts co-authored in this paper. We leveraged the Phenoscape knowledge \cite{edmunds2015phenoscape} base with manual annotations to procure the presence-absence trait matrix. For Bird-10K, we obtained the trait matrix from the attribute annotations provided along with CUB-200-2011. We constructed approximately  $380K$ question-answer pairs for trait identification tasks.

For grounding and referring VQA tasks, the ground truths were manually annotated with the help of expert biologists on our team. We manually annotated bounding boxes corresponding to the traits of 500 fish specimens and 500 bird specimens, which are subsets of the larger Fish-10K and Bird-10K datasets, respectively. We used the CVAT tool \cite{boris_sekachev_2020_4009388} for annotation. The task-specific question formats with the default prompts are provided in Section \ref{sec:prompts}.

\section{Dataset Card}
We provide the dataset card with a detailed description of the metadata, data instances, annotation, and license information here \url{https://huggingface.co/datasets/sammarfy/VLM4Bio#dataset-card-for-vlm4bio}.

\section{Links to Access the Dataset and Its Metadata}
We provide a GitHub link \url{https://github.com/sammarfy/VLM4Bio} and an accessible Hugging Face link \url{https://huggingface.co/datasets/sammarfy/VLM4Bio} to access the dataset and its metadata.

\section{Dataset Availability and Maintanance}
The VLM4Bio dataset and metadata are available in a Hugging Face repository. To access the VLM4Bio dataset, please visit \url{https://huggingface.co/datasets/sammarfy/VLM4Bio}. Long-term support and maintenance of the dataset will be provided by our team. We have published a code repository for dataset preprocessing, including tasks such as downloading the dataset, reading images and metadata, cropping images, and running the evaluation experiments presented in the VLM4Bio paper. To access the VLM4Bio code repository, please visit \url{https://github.com/sammarfy/VLM4Bio}.

\section{Data Licenses}

VLM4Bio dataset is licensed as \href{https://creativecommons.org/licenses/by/4.0/}{Creative Commons Attribution 4.0 International}. The images of the corresponding organisms are licensed as follows:

\begin{enumerate}
    \item Fish Dataset License: \href{https://creativecommons.org/licenses/by-nc/4.0/}{CC BY-NC}.
    \item  Bird Dataset License: \href{https://creativecommons.org/licenses/by/4.0/}{CC BY}.
    \item Butterfly Dataset License: \href{https://creativecommons.org/licenses/by/4.0/}{Creative Commons Attribution 4.0 International}.
\end{enumerate}

We provide image-specific licenses in the dataset card \url{https://huggingface.co/datasets/sammarfy/VLM4Bio#licensing-information}. We have hosted the dataset on HuggingFace, and the DOI will be made publicly available upon publication.

\section{Data Distribution and Key Statistics}

\begin{table*}[t]
\centering
\setlength\tabcolsep{2pt} 
\renewcommand{\arraystretch}{1.5}
\resizebox{\textwidth}{!}{
\begin{tabular}{crrrrrrrrrrrrrrr}
\toprule
\multicolumn{1}{c}{} & \multicolumn{15}{c}{\textbf{Datasets}} \\
\cmidrule(lr){2-16}
\textbf{Statistics} & \textit{Fish-10$K$} & \textit{Bird-10$K$} & \textit{Butterfly-10$K$} & \textit{Fish-500} & \textit{Bird-500} & \renewcommand{\arraystretch}{1.2}{\begin{tabular}[c]{@{}c@{}}\textit{Fish-} \\ \textit{Easy}\end{tabular}} & \renewcommand{\arraystretch}{1.2}{\begin{tabular}[c]{@{}c@{}}\textit{Fish-} \\ \textit{Medium}\end{tabular}} & \renewcommand{\arraystretch}{1.2}{\begin{tabular}[c]{@{}c@{}}\textit{Bird-} \\ \textit{Easy}\end{tabular}} & \renewcommand{\arraystretch}{1.2}{\begin{tabular}[c]{@{}c@{}}\textit{Bird-} \\ \textit{Medium}\end{tabular}} & \renewcommand{\arraystretch}{1.2}{\begin{tabular}[c]{@{}c@{}}\textit{Butterfly-} \\ \textit{Easy}\end{tabular}} & \renewcommand{\arraystretch}{1.2}{\begin{tabular}[c]{@{}c@{}}\textit{Butterfly-} \\ \textit{Medium}\end{tabular}} & \renewcommand{\arraystretch}{1.2}{\begin{tabular}[c]{@{}c@{}}\textit{Butterfly-} \\ \textit{Hard}\end{tabular}} & \renewcommand{\arraystretch}{1.2}{\begin{tabular}[c]{@{}c@{}}\textit{Fish-} \\ \textit{Prompting}\end{tabular}} & \renewcommand{\arraystretch}{1.2}{\begin{tabular}[c]{@{}c@{}}\textit{Bird-} \\ \textit{Prompting}\end{tabular}} & \renewcommand{\arraystretch}{1.2}{\begin{tabular}[c]{@{}c@{}}\textit{Butterfly-} \\ \textit{Prompting}\end{tabular}}\\
\midrule
\textbf{Images}  & 10,347 & 11,092 & 10,013 & 500 & 492 & 200 & 200 & 200 & 200 & 200 & 200 & 200 & 500 & 500 &  500 \\
\textbf{Species} & 495    & 188    & 60     & 60  & 47  & 51 & 10 & 50 & 10 & 50 & 10 & 1 & 25 & 37 & 25  \\
\textbf{Genera}  & 178    & 114    & 27    & 18   & 33  & 10 & 1 & 10 & 1 & 10 & 1 & 1 & 12 & 30 & 10  \\
\textbf{Traits}  & 10     & 28     & -     & 8    & 5   & - & - & - & - & - & - & - & - & - & -  \\
\bottomrule
\end{tabular}}
\caption{Statistics of the VLM4Bio dataset.}
\label{tab:dataset-stat}
\end{table*}

\begin{figure}[t]
\begin{subfigure}{.33\textwidth}
  \centering
  \includegraphics[width=\linewidth]{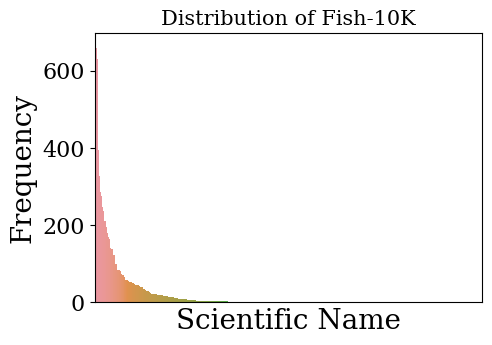}
  \label{fig:dist-fish}
\end{subfigure}
\begin{subfigure}{.33\textwidth}
  \centering
  \includegraphics[width=\linewidth]{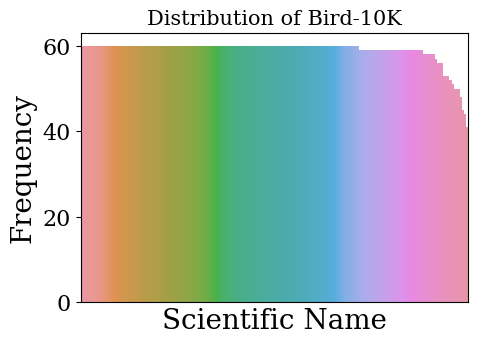}
  \label{fig:dist-bird}
\end{subfigure}
\begin{subfigure}{.33\textwidth}
  \centering
  \includegraphics[width=\linewidth]{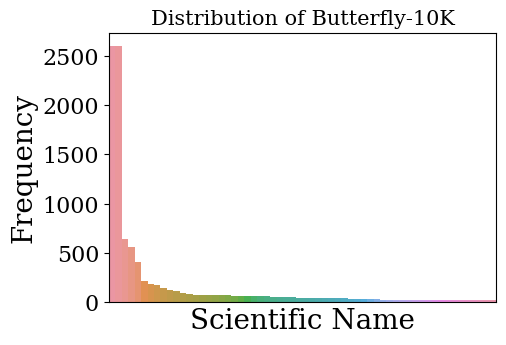}
  \label{fig:dist-butterfly}
\end{subfigure}
\caption{Dataset Distribution of Fish-$10K$, Bird-$10K$, and Butterfly-$10K$.}
\label{fig:distributions}
\end{figure}

Table \ref{tab:dataset-stat} provides the key statistics for the datasets, including the number of images, species, genera, and traits present in each one. We are examining the Zero-shot accuracy of the VLMs on Fish-10K, Bird-10K, and Butterfly-10K for Species Classification and Trait Identification tasks, Fish-500 and Bird-500 for Trait Grounding, Trait Referring and Trait Counting, and easy, medium, hard, prompting datasets for analyzing the role of answer choices, VLM reasoning and hallucination tests. From Figure \ref{fig:distributions}, it is clear that Fish-10K and Butterfly-10K are imbalanced, with a bias toward some species that are more common in our environment (such as \textit{Heliconius erato} and \textit{Heliconius melpomene} for Butterflies). The imbalance in Fish-10K and Butterfly-10K reflects the natural imbalance in the occurrence and observation of species in museum collections. Due to the scarcity of images for the rare species, it is difficult to increase their representation to avoid imbalance. As a result, we have included many under-represented species in the Fish and Butterfly datasets to report performance on the rare classes. In contrast, the Bird-10K dataset is well-balanced, with most species having 60 images. The easy, medium, hard, and prompting datasets are also balanced, which ensures a comprehensive evaluation of the zero-shot performance of the competing VLMs.

\section{Traits Considered for the Task of Trait Identification}
\begin{figure*}[ht]
    \centering
    \includegraphics[width=\linewidth]{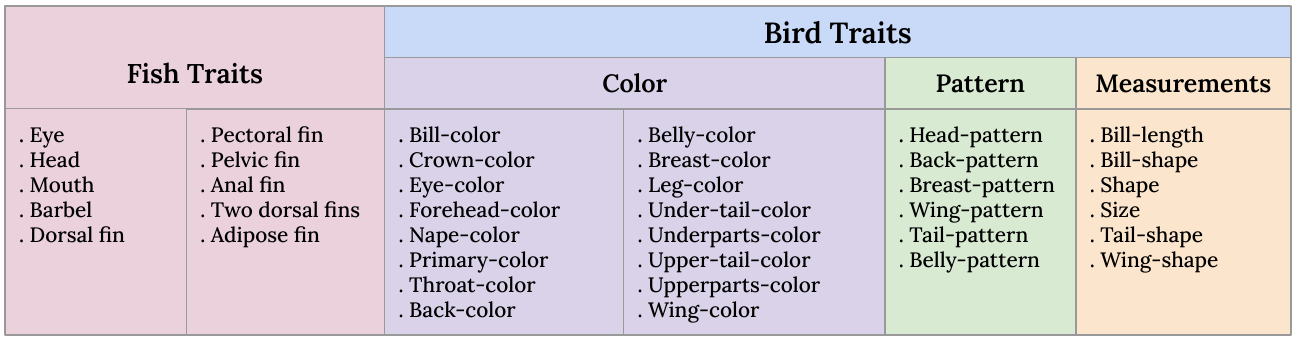}
    \caption{Trait list for Trait Identification task.}
    \label{fig:identification-traits}
\end{figure*}
Figure \ref{fig:identification-traits} shows the Fish traits and Bird traits used for evaluating the VLM's performance in the identification task.
For fishes, we considered $10$ binary (presence/absence) traits which include the \textit{eye, head, mouth, barrel, dorsal fin, pectoral fin, pelvic fin, anal fin, and adipose fin}. We generated MC questions for the presence of each trait in an image (with two options: yes or no). Whereas for birds, we considered $28$ traits covering their color, pattern, and measurements (size and shape of regions) in a multiple-choice format.

\section{Traits Considered for the Tasks of Trait Grounding and Referring}
To evaluate the VLM performance in Grounding and Referring, we identified $8$ traits for fish and $5$ traits for birds. Specifically, we manually annotated the \textit{dorsal fin, adipose fin, caudal fin, anal fin, pelvic fin, pectoral fin, head, and eye} of the $500$ fish specimens. Similarly, for birds, we annotated the \textit{beak, head, eye, wings, and tail}. Trait grounding and referring tasks are carried out using the Fish-500 and Bird-500 datasets.

\section{VLM Baselines}
 We consider the following VLM baselines to evaluate the performance on VLM4Bio dataset: (1) GPT-4V(ision) \cite{openai2023gpt}, which is a proprietary VLM from OpenAI, that uses a generative pre-trained transformer model capable of understanding and generating both text and visual contents, (2) LLaVA-v1.5 (7B/13B) \cite{liu2023visual}, which builds on top of the Vicuna LLM \cite{chiang2023vicuna} by linearly projecting the visual embedding into the word embedding space. The LLaVA model has two different variants with 7B and 13B parameters, respectively, that depend on the size of the base Vicuna model,  (3) COG-VLM \cite{wang2023cogvlm}, which performs a simple concatenation of the image and the text modalities, and uses trainable visual layers in the text-based transformer blocks, (4) MiniGPT-4 (Vicuna 7B/13B) \cite{zhu2023minigpt}, which is similar to LLaVA as it is built on top of the Vicuna model and linearly projects the visual embeddings for better understanding. Similar to LLaVA, MiniGPT-4 is available in two variants depending on the type the base Vicuna model (Vicuna 7B/13B), (5) BLIP-FLAN-T5-XL/XXL \cite{li2023blip}, which utilizes an effective pre-training strategy that relies on bootstrapping from frozen-pretrained CLIP encoders and LLMS by using a querying transformer block (available as two variants: XL and XXL), and (6) Instruct-BLIP (Vicuna 7B/13B) \cite{dai2023instructblip}, which performs finetuning on BLIP-2 with visual-instruction tuning data to improve zero-shot capabilities of BLIP-2 (available as two variants depending on the Vicuna model: Vicuna 7B/13B).

\section{Prompts to Evaluate VLM performance}
\begin{figure*}[t]
    \centering
    \includegraphics[width=\linewidth]{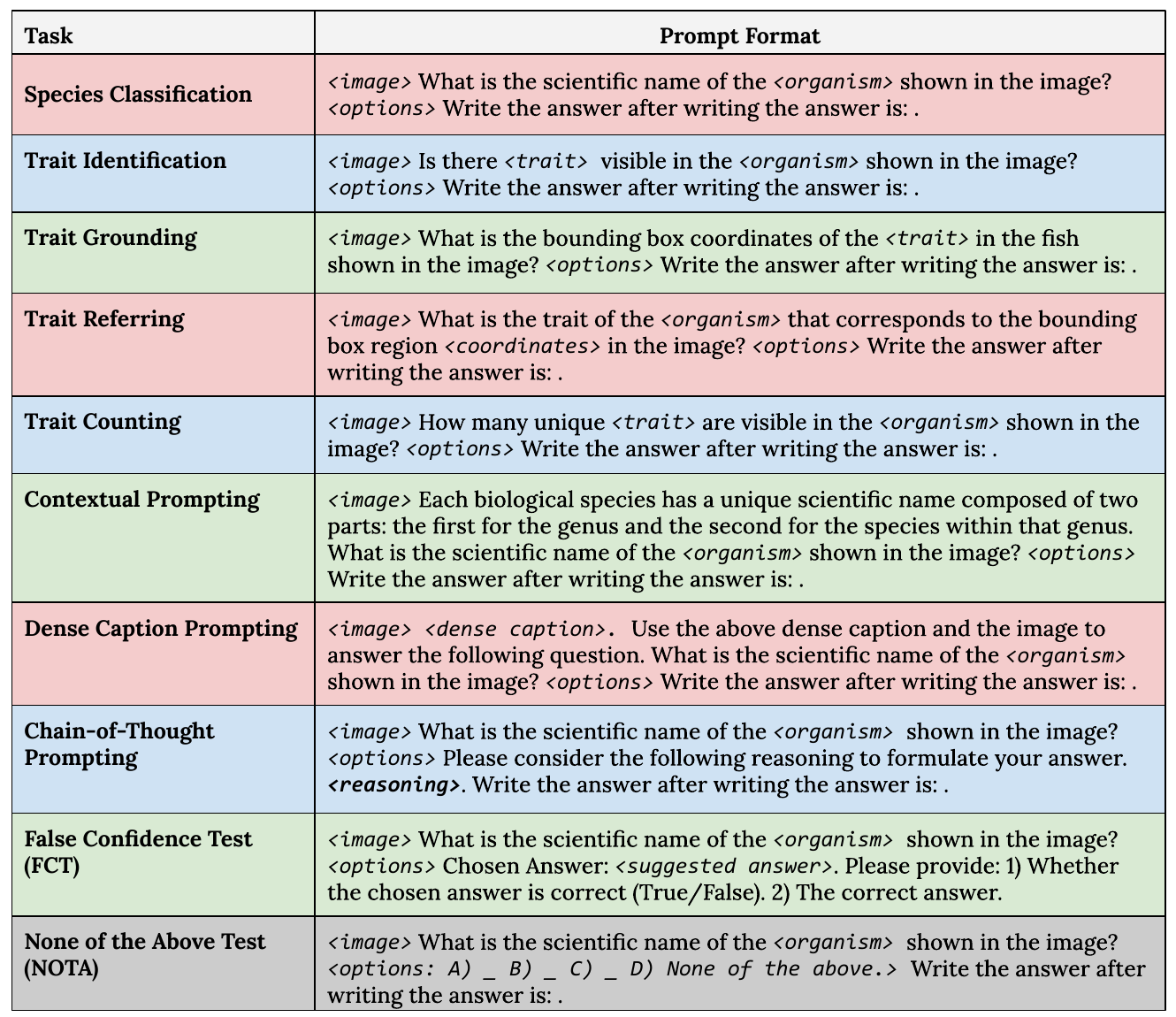}
    \caption{Prompts Templates used for Evaluation. There will be no \texttt{<options>} for Open set questions.}
    \label{fig:prompts}
\end{figure*}
In order to ensure a fair comparison of the VLM responses to different types of questions in our dataset, we used the same question prompt for all the models across the various scientific tasks. It's worth noting that each model may perform differently with different prompts. However, for the sake of simplicity in our evaluation, we opted for a consistent prompt for all the models. The prompts specific to each task are displayed in Figure \ref{fig:prompts}.

\section{Error Analyses for VLM Responses}
\label{sec:prompts}

\begin{figure}[ht]
\begin{subfigure}{.5\textwidth}
  \centering
  \includegraphics[width=\linewidth]{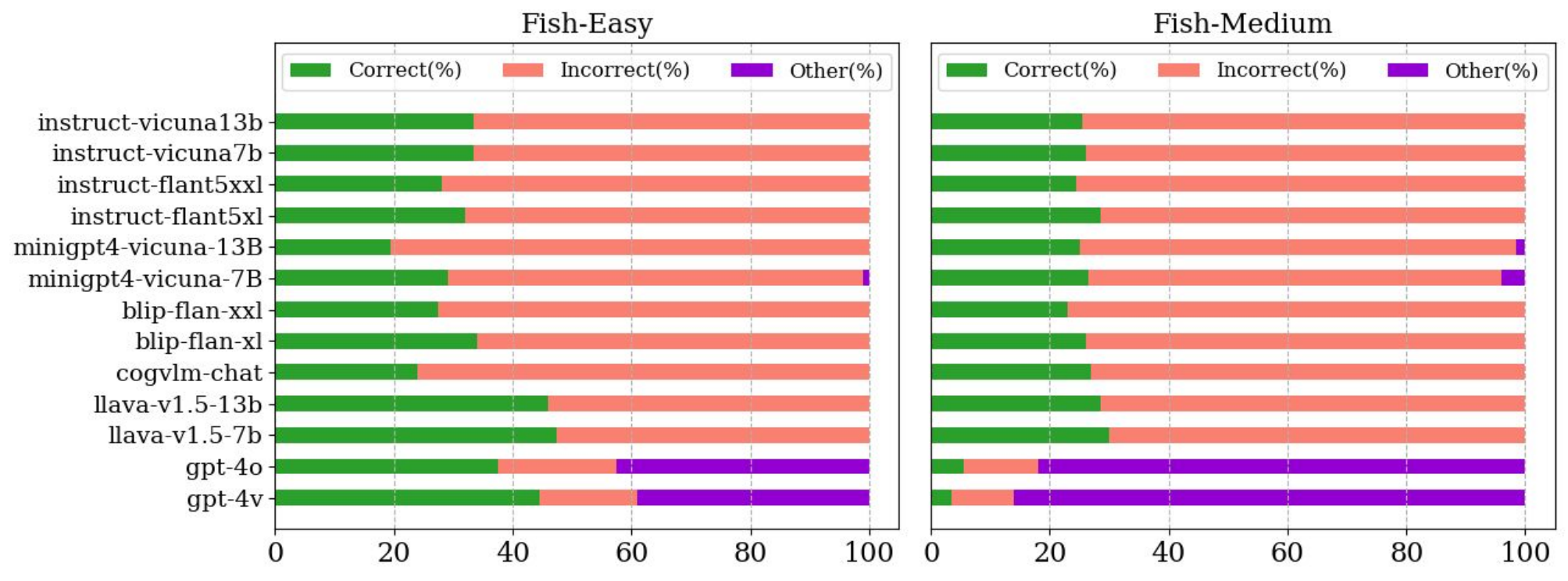}
  \caption{Error Analysis for Fish-Easy and -Medium.}
  \label{fig:cls-err-fish}
\end{subfigure}
\begin{subfigure}{.5\textwidth}
  \centering
  \includegraphics[width=\linewidth]{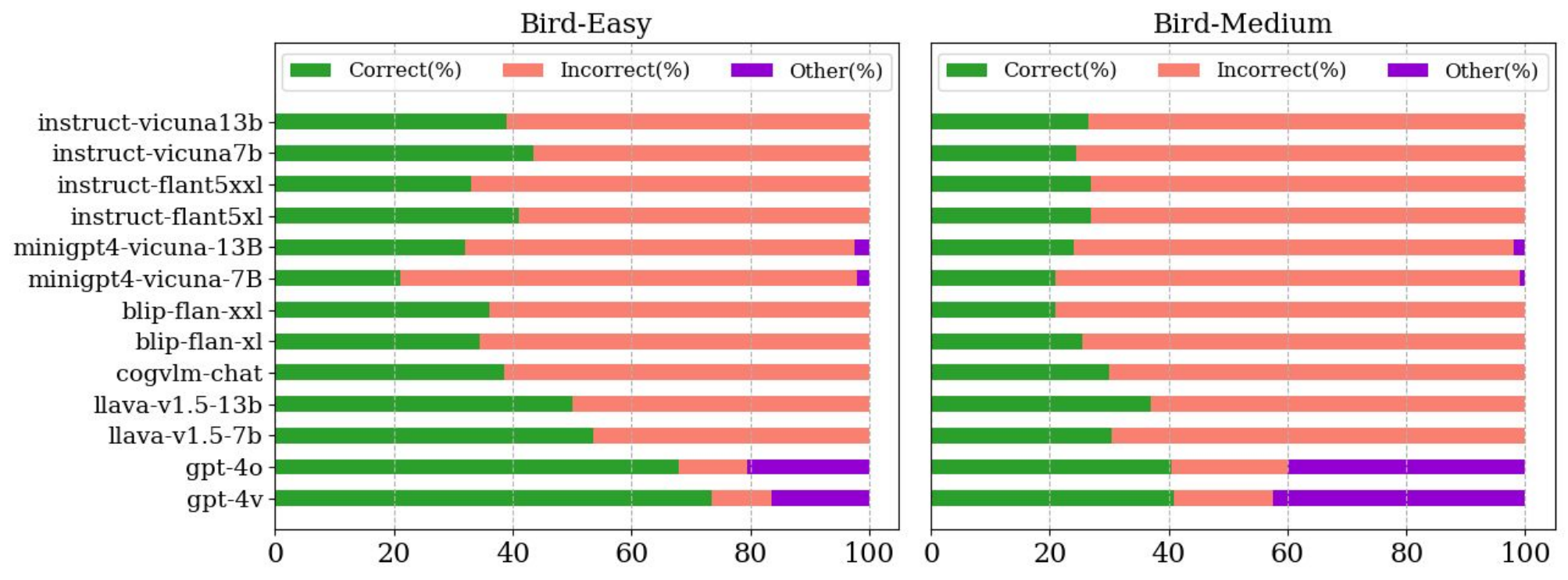}
  \caption{Error Analysis for Bird-Easy and -Medium.}
  \label{fig:cls-err-bird}
\end{subfigure}
\begin{subfigure}{.5\textwidth}
  \centering
  \includegraphics[width=\linewidth]{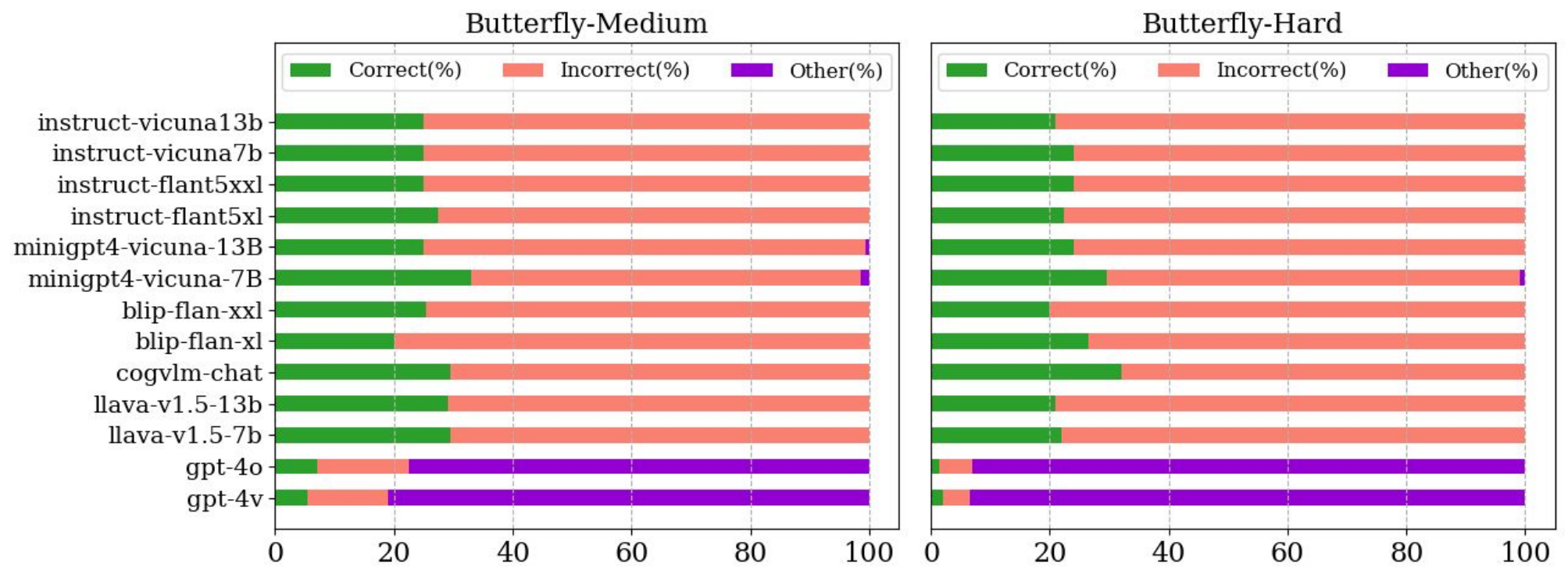}
  \caption{Error Analysis for Butterfly-Medium and -Hard.}
  \label{fig:cls-err-butterfly}
\end{subfigure}
\begin{subfigure}{.38\textwidth}
  \centering
  \includegraphics[width=\linewidth]{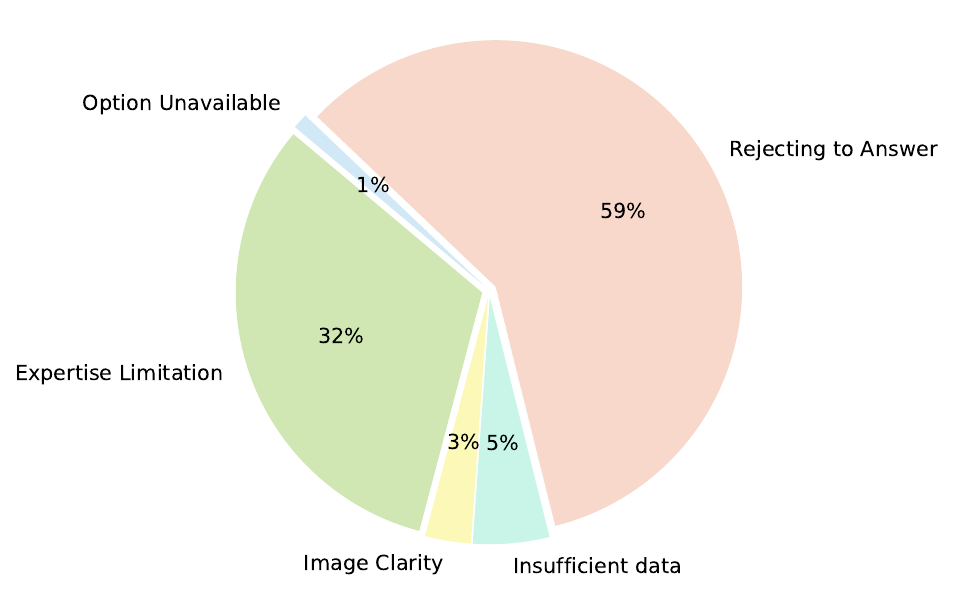}
  \caption{Categories for 250 annotated GPT-4V ``Other'' responses.}
  \label{fig:gpt_pie_chart}
\end{subfigure}
\caption{Analysis of errors for the pretrained VLM responses.}
\end{figure}


We categorize the VLM responses into 3 categories: (1) \emph{Correct (\%)}: where the scientific name is accurately predicted, (2)  \emph{Incorrect (\%)}: where the scientific name is incorrect, and (3) \emph{Other (\%)}: a special category for instances where the model abstains from providing a scientific name. 

Figure \ref{fig:cls-err-fish}, \ref{fig:cls-err-bird} and \ref{fig:cls-err-butterfly} show the distribution of errors of different VLMs on Fish-Easy and Fish-Medium, Bird-Easy and Bird-Medium, and Butterfly-Medium and Butterfly-Hard datasets respectively using stacked-bar plots showing the three categories of VLM predictions. GPT-4V, for instance, shows a reduced rate of incorrect responses but a higher incidence of ``Other'' responses for these datasets, which include apologetic expressions, admissions of inability to precisely visualize the organism, and disclaimers regarding prediction without sufficient expert data and guidance.

To further analyze the type of errors happening in the other (\%) category of VLM predictions, we manually examined 250 randomly selected ``Other'' GPT-4V responses for the task of fish species classification (MC question type) to generate the pie-chart of error categories shown in Figure \ref{fig:gpt_pie_chart}. We can see that a majority of the ``Other'' responses belong to the category: \textit{Rejecting to Answer} (59\%), where the GPT-4V states that it is unable to provide an answer, sometimes stating the reason that it cannot answer based on a single image. We also observe a large fraction of \textit{Expertise Limitation} responses where GPT-4V states that an expert taxonomist is needed to answer the question and its capabilities do not include recognizing or confirming species based on visual data. The next major type of ``Other'' responses are \textit{Insufficient Data}, where GPT-4V states that it requires additional data to answer the question, e.g., taxonomic information or habitat information. The other error categories include \textit{Image Clarity} issues and \textit{Option Unavailable} (i.e., GPT-4V could not find a suitable option from the list of options provided in the prompt).

\section{Comparing Pre-trained VLMs with a Biologically Fine-tuned Model}

\begin{table*}[ht]
\centering
\setlength\tabcolsep{2pt} 
\fontsize{8pt}{10}\selectfont
\begin{tabular}{ccrrrrr}
\toprule
\multicolumn{2}{c}{} & \multicolumn{5}{c}{\textbf{Models}} \\
\cmidrule(lr){3-7}
\textbf{Dataset }& \begin{tabular}[c]{@{}c@{}}\textbf{Question} \\ \textbf{type}\end{tabular}  & \textit{gpt-4v} &  \begin{tabular}[c]{@{}c@{}}\textit{llava} \\ \textit{v1.5-7b}\end{tabular} &  \begin{tabular}[c]{@{}c@{}}\textit{cogvlm} \\ \textit{chat}\end{tabular} & \begin{tabular}[c]{@{}c@{}}\textit{CLIP}\end{tabular} &   \textit{BioCLIP} \\
\midrule
\multicolumn{7}{c}{\textbf{Species Classification}}\\
\midrule
\textbf{Fish-10K} & Open & 1.01 &\cellcolor{blue!15}2.32 & \cellcolor{red!15}0.11 & 0.57 &  1.24 \\
& MC & 35.91 & 40.20 & \cellcolor{red!15}31.72 & 42.45 &  \cellcolor{blue!15}50.65 \\
\midrule
\textbf{Bird-10K} & Open & 17.40 & 1.45 & \cellcolor{red!15}0.86 & 7.74 & \cellcolor{blue!15}67.12 \\
& MC & 82.58 & 50.32 & \cellcolor{red!15}44.73  & 45.78 &  \cellcolor{blue!15}93.93 \\
\midrule
\textbf{Butterfly-10K} & Open & 0.04 & 0.05 & \cellcolor{red!15}0.01 & 5.33 &  \cellcolor{blue!15}15.95 \\
& MC & \cellcolor{red!15}28.91 & 50.24 &  36.45 & 45.60 & \cellcolor{blue!15}62.32 \\
\bottomrule
\end{tabular}
\caption{Zero-shot accuracy comparison of VLM baselines (in \% ranging from 0 to 100) with BioCLIP for the species classification task.
Results are color-coded as \colorbox{blue!15} {Best}, and \colorbox{red!15} {Worst}.}
\label{tab:bioclip}
\end{table*}

We compare the large pretrained VLMs and BioCLIP \cite{stevens2023bioclip}, a state-of-the-art foundation model for species classification. Furthermore, we include the simple CLIP model pretrained with OpenAI weights \cite{radford2021learning} to evaluate the zero-shot classification performance. Our evaluation was carried out on the Fish-10K, Bird-10K, and Butterfly-10K datasets, and the results are presented in Table \ref{tab:bioclip}. We can see that BioCLIP significantly outperforms large pretrained VLMs on the Bird-10K and Butterfly-10K datasets, suggesting that BioCLIP may have been trained on images that are similar to the organisms present in these datasets. However, as noted in the paper, BioCLIP is not trained on fish images, and hence, the performance of large VLMs is similar to that of BioCLIP on Fish-10K images. We can also see that despite BioCLIP's ability to effectively select the correct scientific name from a smaller set of options in multiple-choice (MC) questions, its performance significantly declines when asked to choose the scientific name from a larger set of open questions. From our observation, it is noteworthy that fine-tuning biological images with scientific names can help improve the overall accuracy of species classification, suggesting directions for future research in this area.


\begin{figure}[ht]
    \centering
    \begin{subfigure}[b]{0.24\textwidth}
        \centering
        \includegraphics[width=\textwidth]{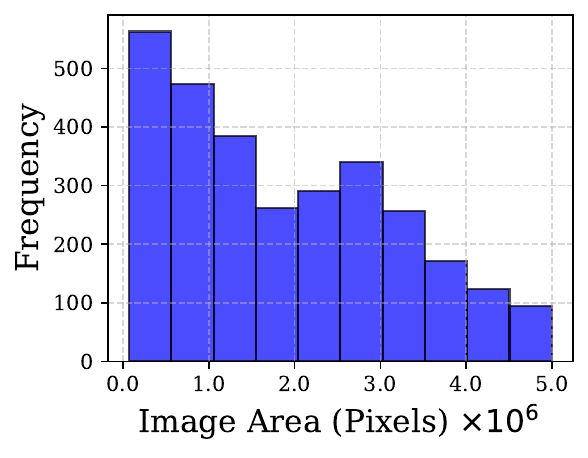}
        \caption{Fish-$10K$}
        \label{fig:fish_resolution}
    \end{subfigure}
    \begin{subfigure}[b]{0.24\textwidth}
        \centering
        \includegraphics[width=\textwidth]{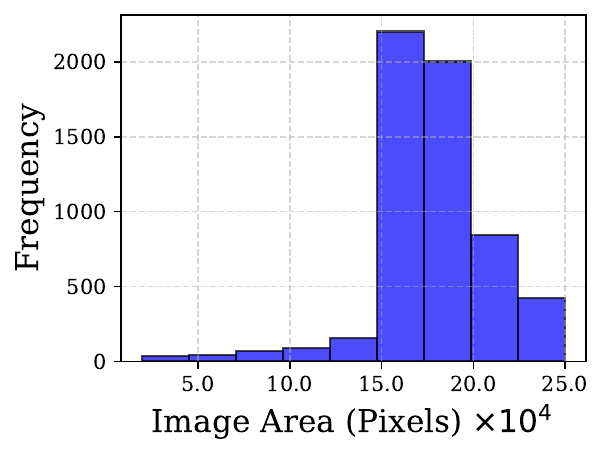}
        \caption{Bird-$10K$}
        \label{fig:bird_resolution}
    \end{subfigure}
    \begin{subfigure}[b]{0.24\textwidth}
        \centering
        \includegraphics[width=\textwidth]{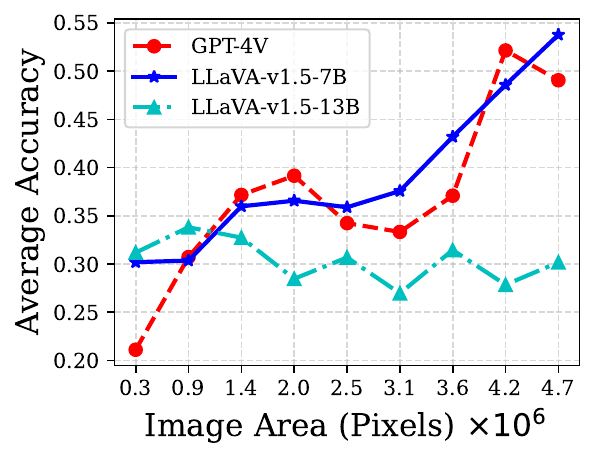}
        \caption{Fish-$10K$}
        \label{fig:fish_performance_vs_resolution}
    \end{subfigure}
    \begin{subfigure}[b]{0.24\textwidth}
        \centering
        \includegraphics[width=\textwidth]{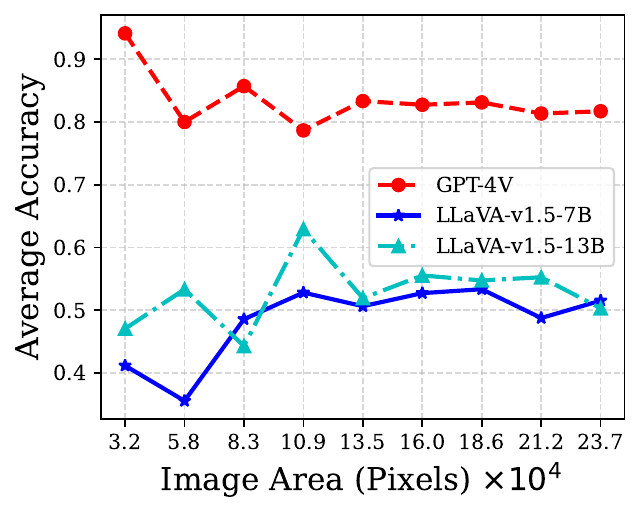}
        \caption{Bird-$10K$}
        \label{fig:bird_performance_vs_resolution}
    \end{subfigure}
    \caption{Distribution of image resolutions for Fish-10K and Bird-10K are shown in Figures (a) and (b), respectively. The average score over image resolution for the GPT-4V, LLaVA-v1.5-7B, and LLaVA-v1.5-13B models on Fish-10K and Bird-10K are presented in Figures (c) and (d). We conduct the experiment in the context of the Species Classification task with Multiple-Choice (MC) questions.}
    \label{fig:resolution}
\end{figure}

\section{Analyzing Effects of Image Resolution on VLM Performance}

To investigate the effect of image resolution on VLM performance, we perform additional experiments summarized in Figure \ref{fig:resolution} of the attached pdf. In this Figure, we show distribution plots for the Fish-10K and Bird-10K datasets with variations in the image resolutions and their impact on the species classification performance (MC question format) for GPT-4V, LLaVA-1.5-7B, and LLaVA-1.5-13B. All the images of the Butterfly-10K have the exact resolution $(500\times333)$; hence, they were not included in the experiment. 
From Figure \ref{fig:fish_performance_vs_resolution}, it is clear that image resolution is influential on the VLM performance for the Fish-10K dataset since higher resolution helps in recognizing the details of the biological traits and correct species. However, for Figure \ref{fig:bird_performance_vs_resolution}, the VLM performances do not vary significantly with the image resolution for the Bird-10K dataset. A potential reason is that the bird dataset is a subset of the CUB dataset, and we hypothesize that the pre-trained VLMs may have seen images with resolutions similar to those in the Bird-10K dataset during training, leading to this behavior.


\section{Case Studies for Effects of Prompting on VLM Performance}

\subsection{No Prompting}
\label{sec:no_prompting}
\begin{figure}[t]
    \centering
    \includegraphics[width=\textwidth]{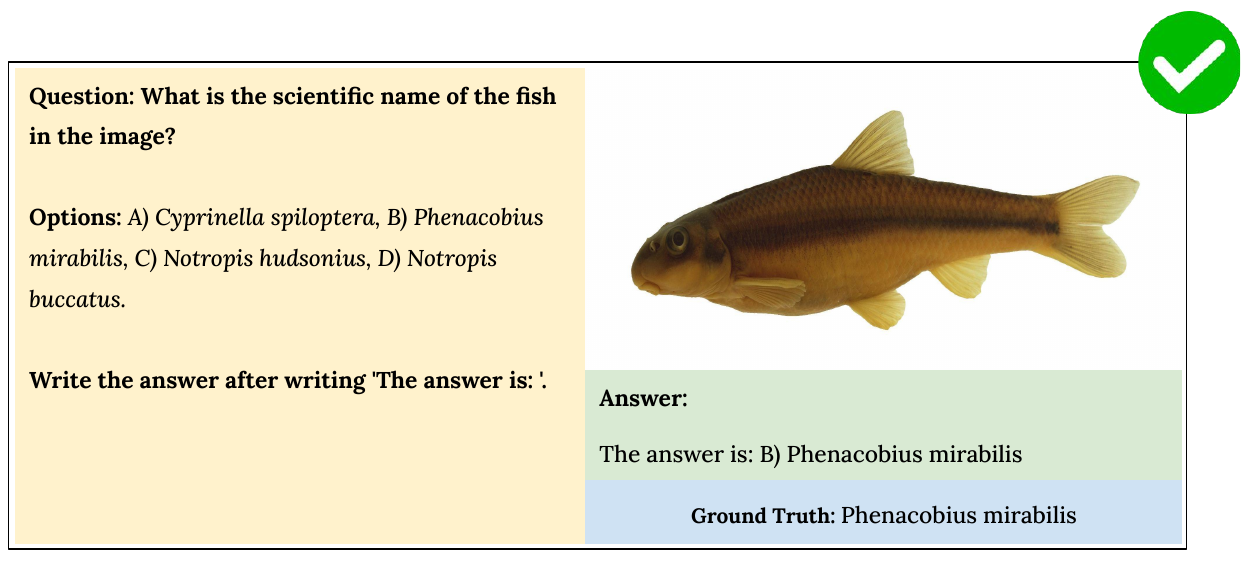}
    \caption{No Prompting. GPT-4o Correct prediction. Section \ref{sec:no_prompting}.}
    \label{gpt_correct_no_prompt_fish}
\end{figure}

\begin{figure}[t]
    \centering
    \includegraphics[width=\textwidth]{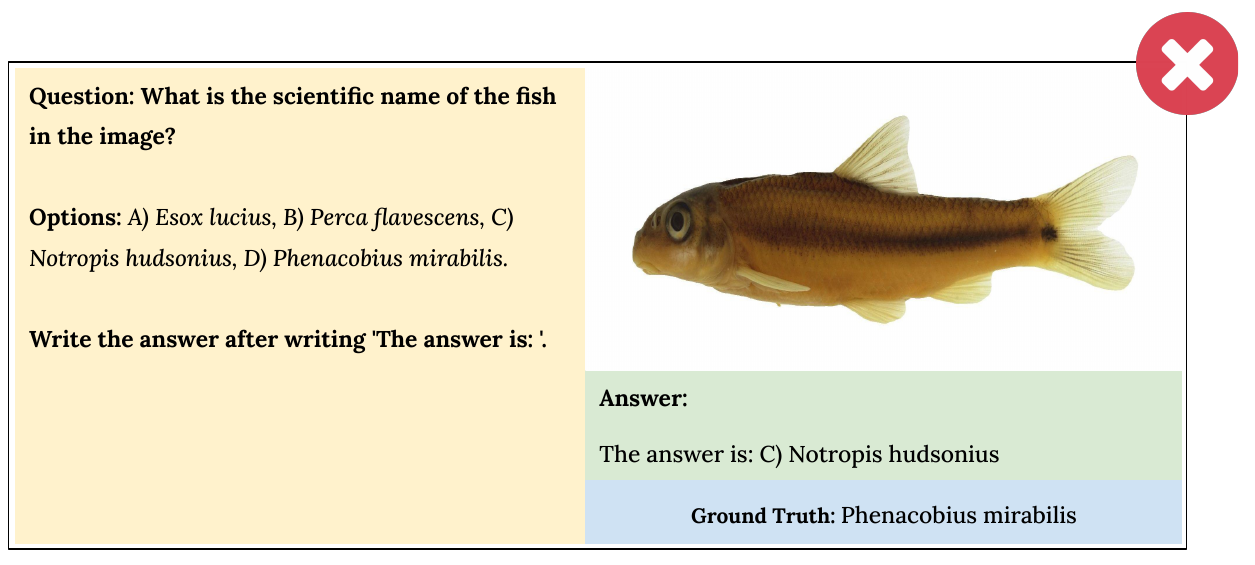}
    \caption{No Prompting. GPT-4o Incorrect prediction. Section \ref{sec:no_prompting}.}
    \label{gpt_incorrect_no_prompt_fish}
\end{figure}

\begin{figure}[t]
    \centering
    \includegraphics[width=\textwidth]{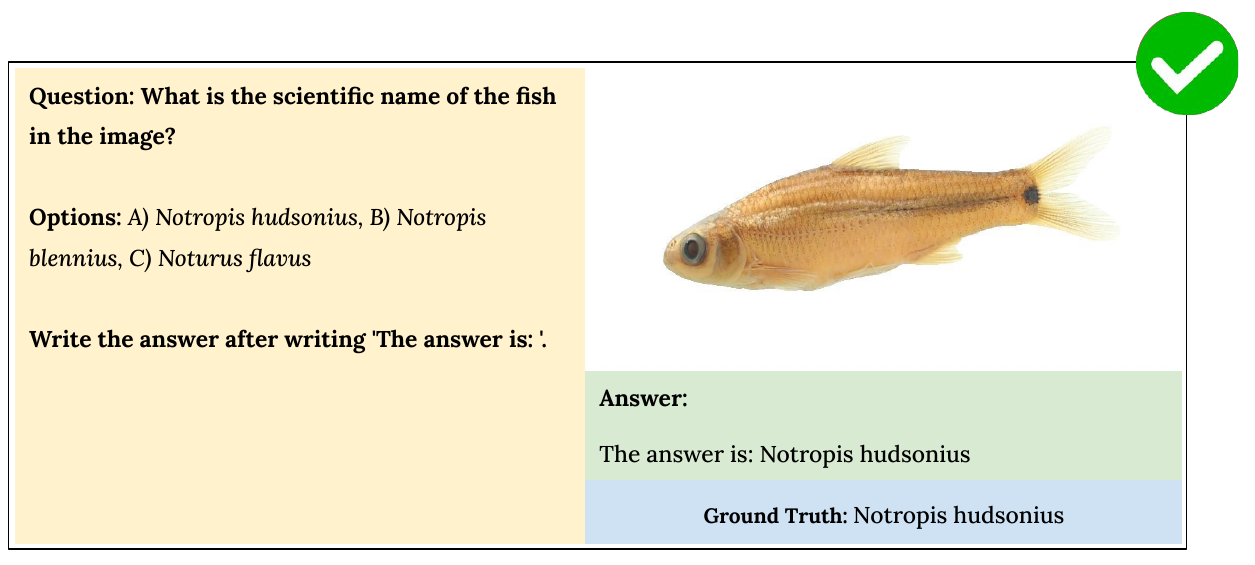}
    \caption{No Prompting. COG-VLM Correct prediction. Section \ref{sec:no_prompting}.}
    \label{cog_correct_no_prompt_fish}
\end{figure}

\begin{figure}[t]
    \centering
    \includegraphics[width=\textwidth]{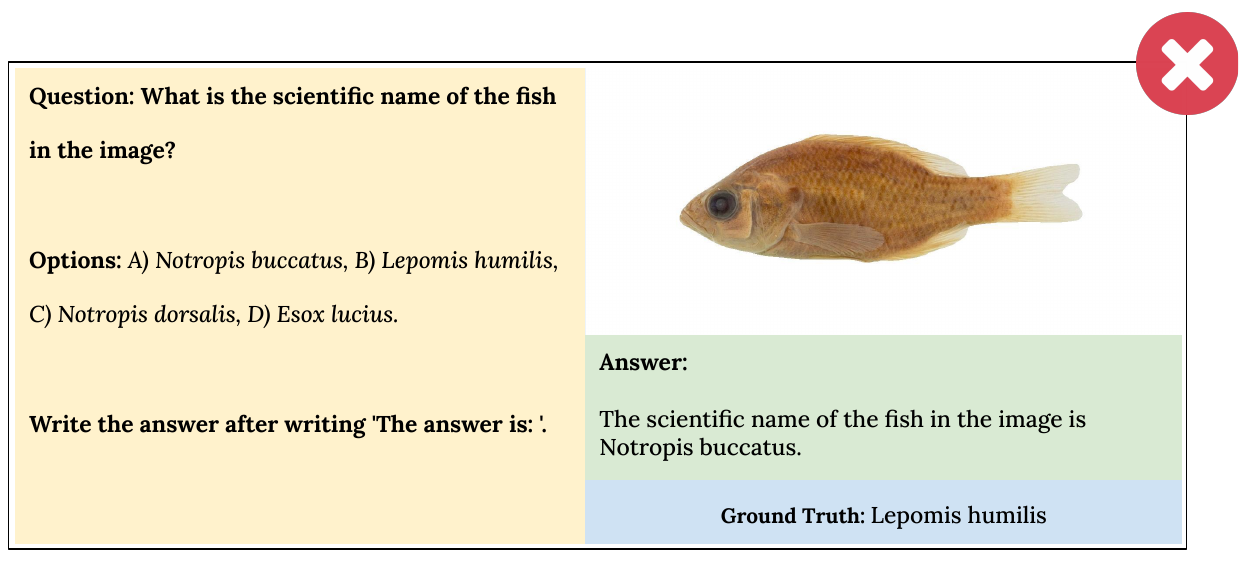}
    \caption{No Prompting. COG-VLM Incorrect prediction. Section \ref{sec:no_prompting}.}
    \label{cog_incorrect_no_prompt_fish}
\end{figure}

\begin{enumerate}
    \item No Prompting. GPT-4o Correct prediction. Refer to Figure \ref{gpt_correct_no_prompt_fish}.
    \item No Prompting. GPT-4o Incorrect prediction. Refer to Figure \ref{gpt_incorrect_no_prompt_fish}.
    \item No Prompting. COG-VLM Correct prediction. Refer to Figure \ref{cog_correct_no_prompt_fish}.
    \item No Prompting. COG-VLM Incorrect prediction. Refer to Figure \ref{cog_incorrect_no_prompt_fish}.
\end{enumerate}

\subsection{Contextual Prompting}
\label{sec:contextual}
\begin{figure}[t]
    \centering
    \includegraphics[width=\textwidth]{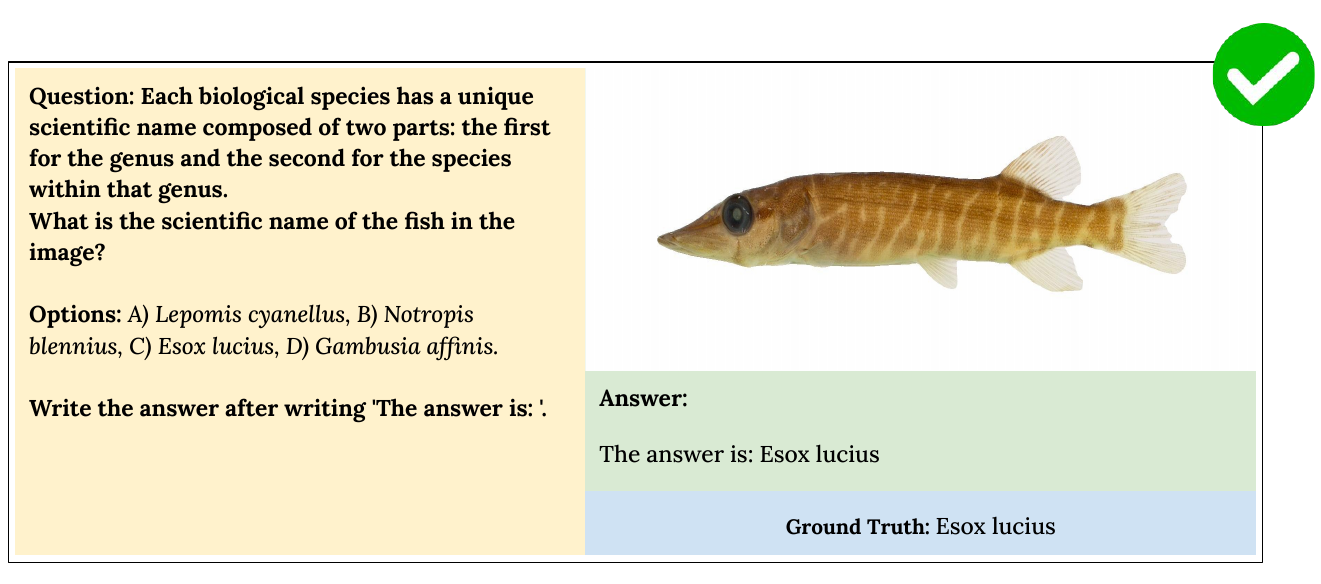}
    \caption{Contextual Prompting. GPT-4o Correct prediction. Section \ref{sec:contextual}.}
    \label{gpt_correct_contextual_fish}
\end{figure}

\begin{figure}[t]
    \centering
    \includegraphics[width=\textwidth]{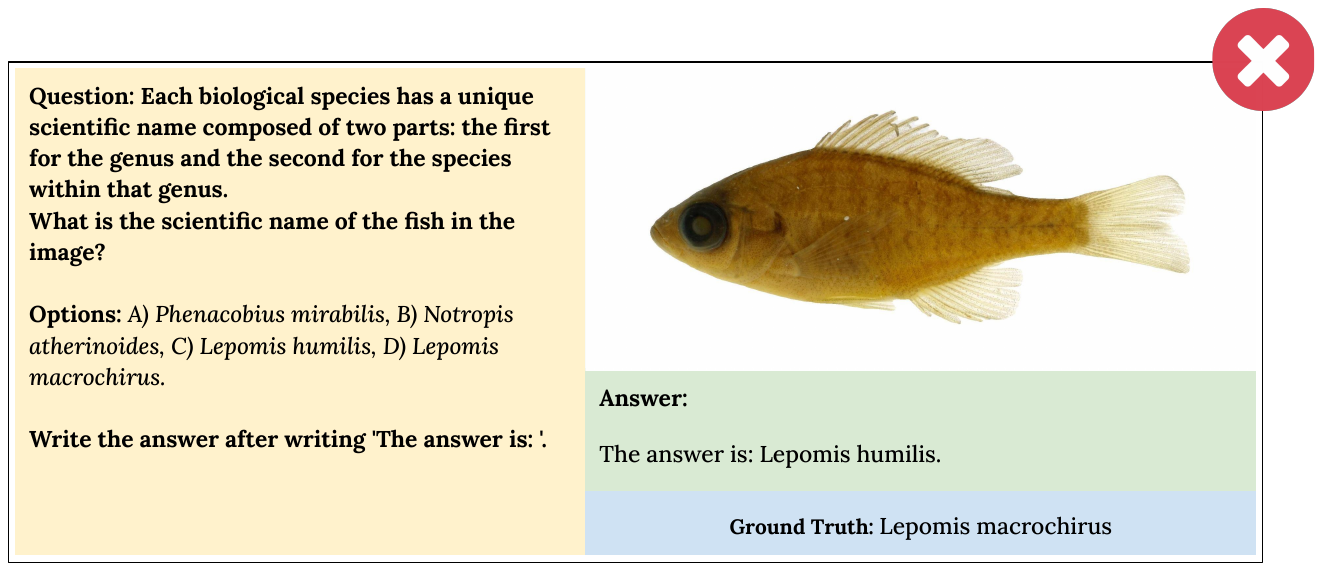}
    \caption{Contextual Prompting. GPT-4o Incorrect prediction. Section \ref{sec:contextual}.}
    \label{gpt_incorrect_contextual_fish}
\end{figure}

\begin{figure}[t]
    \centering
    \includegraphics[width=\textwidth]{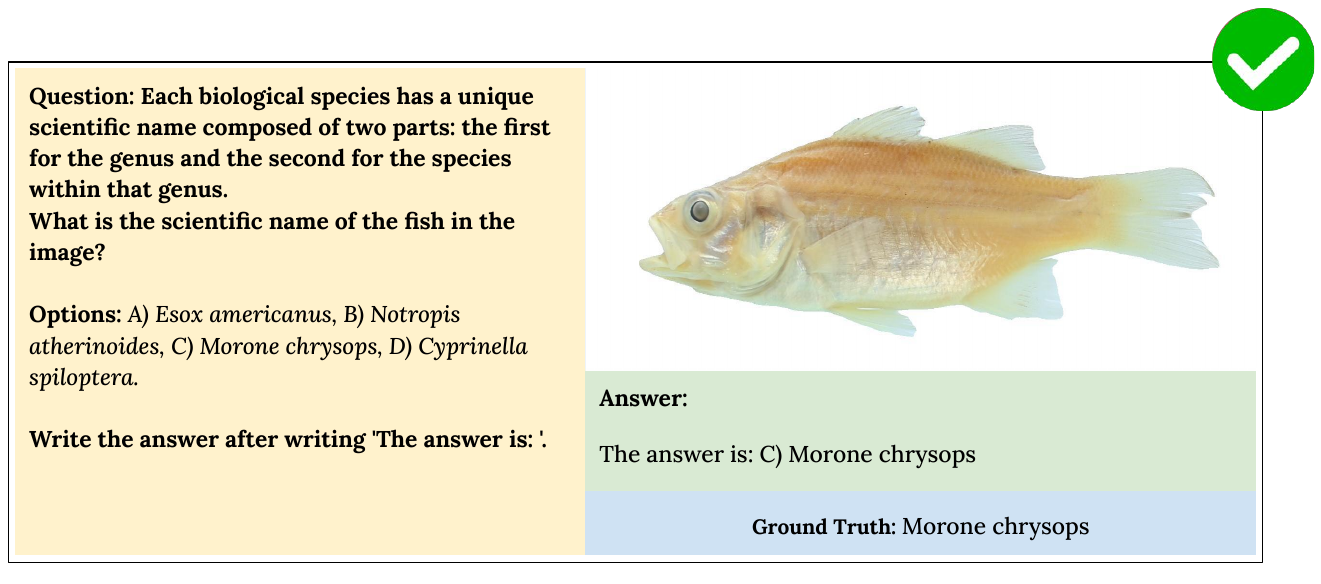}
    \caption{Contextual Prompting. LLaVa-13B Correct prediction. Section \ref{sec:contextual}.}
    \label{llava13b_correct_contextual_fish}
\end{figure}

\begin{figure}[t]
    \centering
    \includegraphics[width=\textwidth]{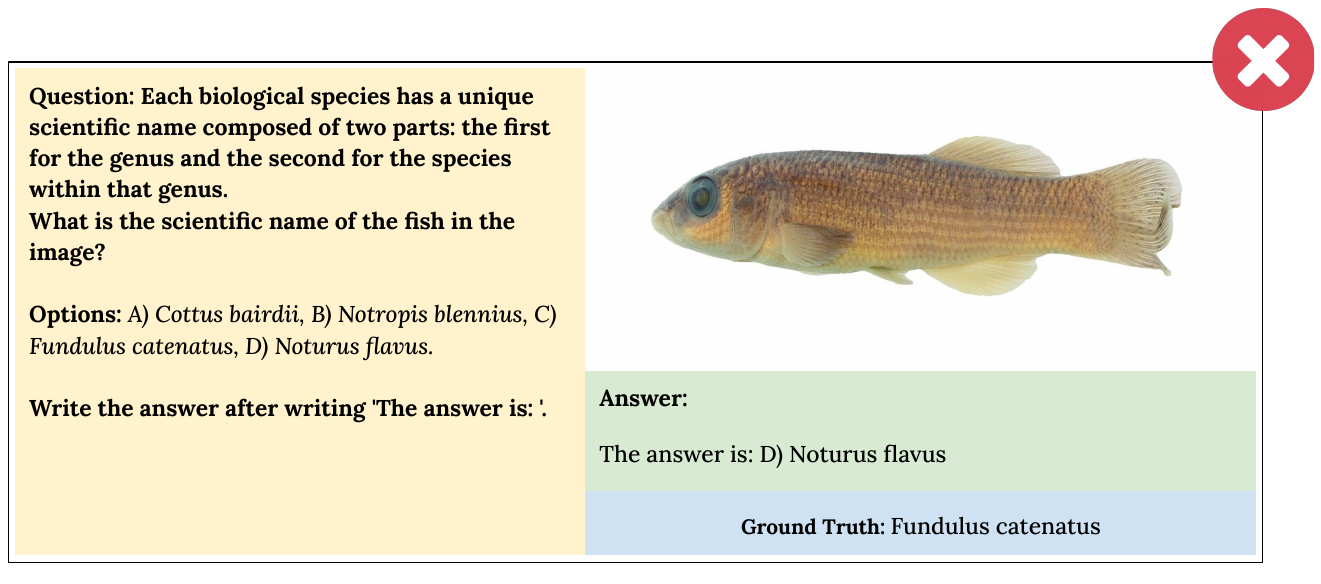}
    \caption{Contextual Prompting. LLaVa-13B Incorrect prediction. Section \ref{sec:contextual}.}
    \label{llava13b_incorrect_contextual_fish}
\end{figure}

\begin{enumerate}
    \item Contextual Prompting. GPT-4o Correct prediction. Refer to Figure \ref{gpt_correct_contextual_fish}.
    \item Contextual Prompting. GPT-4o Incorrect prediction. Refer to Figure \ref{gpt_incorrect_contextual_fish}.
    \item Contextual Prompting. LLaVa-13B Correct prediction. Refer to Figure \ref{llava13b_correct_contextual_fish}.
    \item Contextual Prompting. LLaVa-13B Incorrect prediction. Refer to Figure \ref{llava13b_incorrect_contextual_fish}.
\end{enumerate}

\subsection{Dense Caption}

\label{sec:dense_caption}
\begin{figure}[ht]
    \centering
    \includegraphics[width=\textwidth]{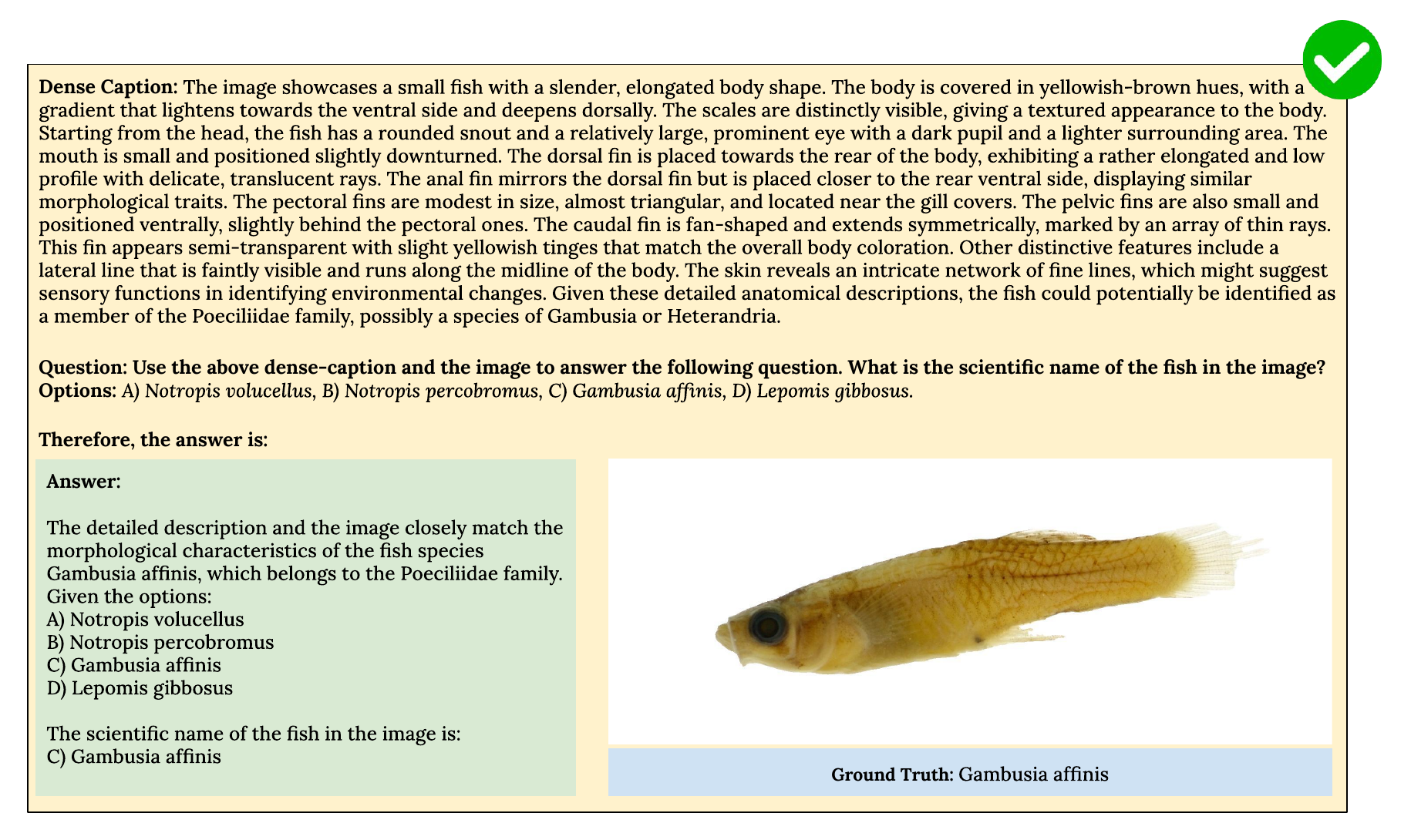}
    \caption{Dense Captions in Prompts. GPT-4o Correct prediction. Section \ref{sec:dense_caption}.}
    \label{gpt_correct_dense_cap_fish}
\end{figure}

\begin{figure}[ht]
    \centering
    \includegraphics[width=\textwidth]{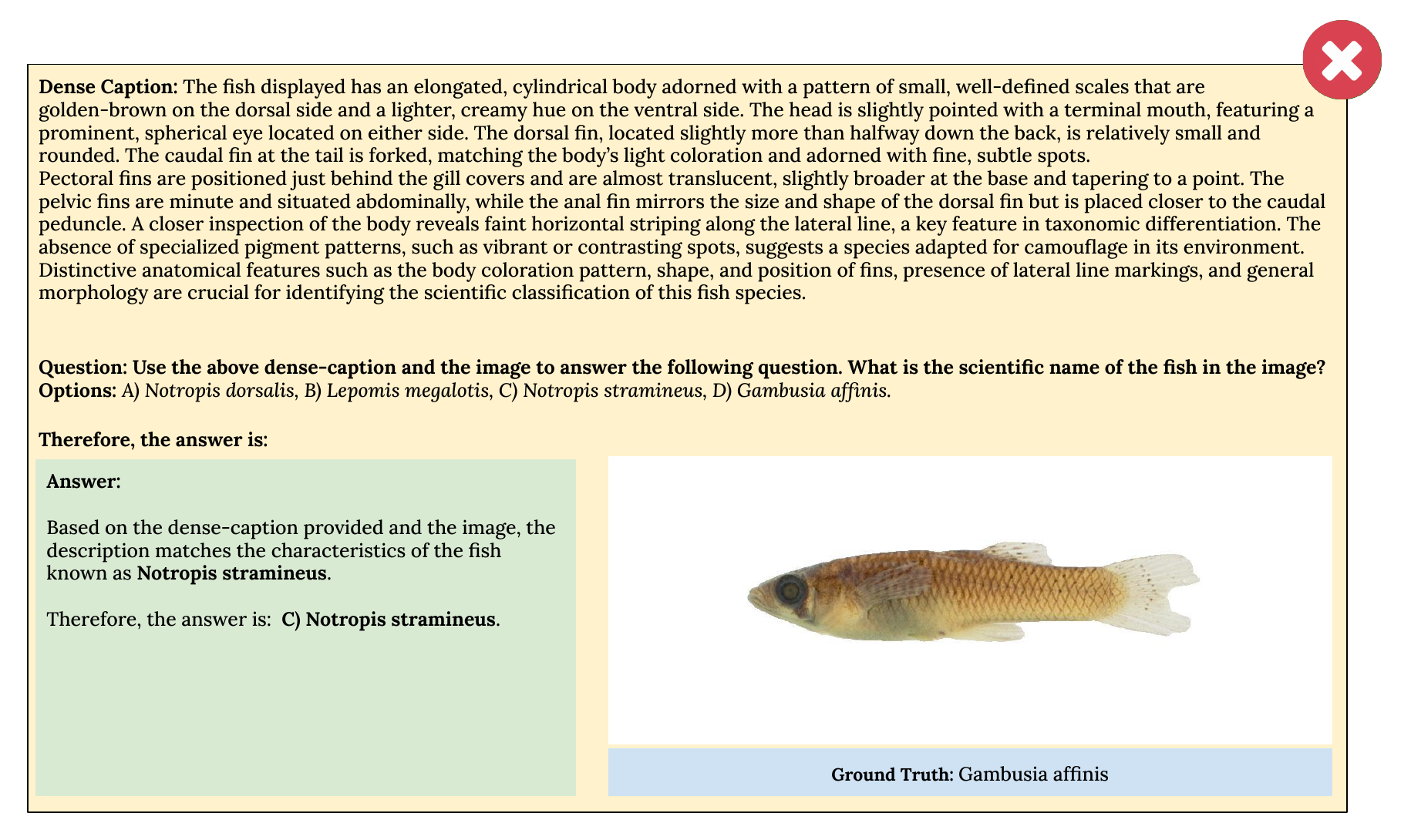}
    \caption{Dense Captions in Prompts. GPT-4o Incorrect prediction. Section \ref{sec:dense_caption}.}
    \label{gpt_incorrect_dense_cap_fish}
\end{figure}

\begin{figure}[ht]
    \centering
    \includegraphics[width=\textwidth]{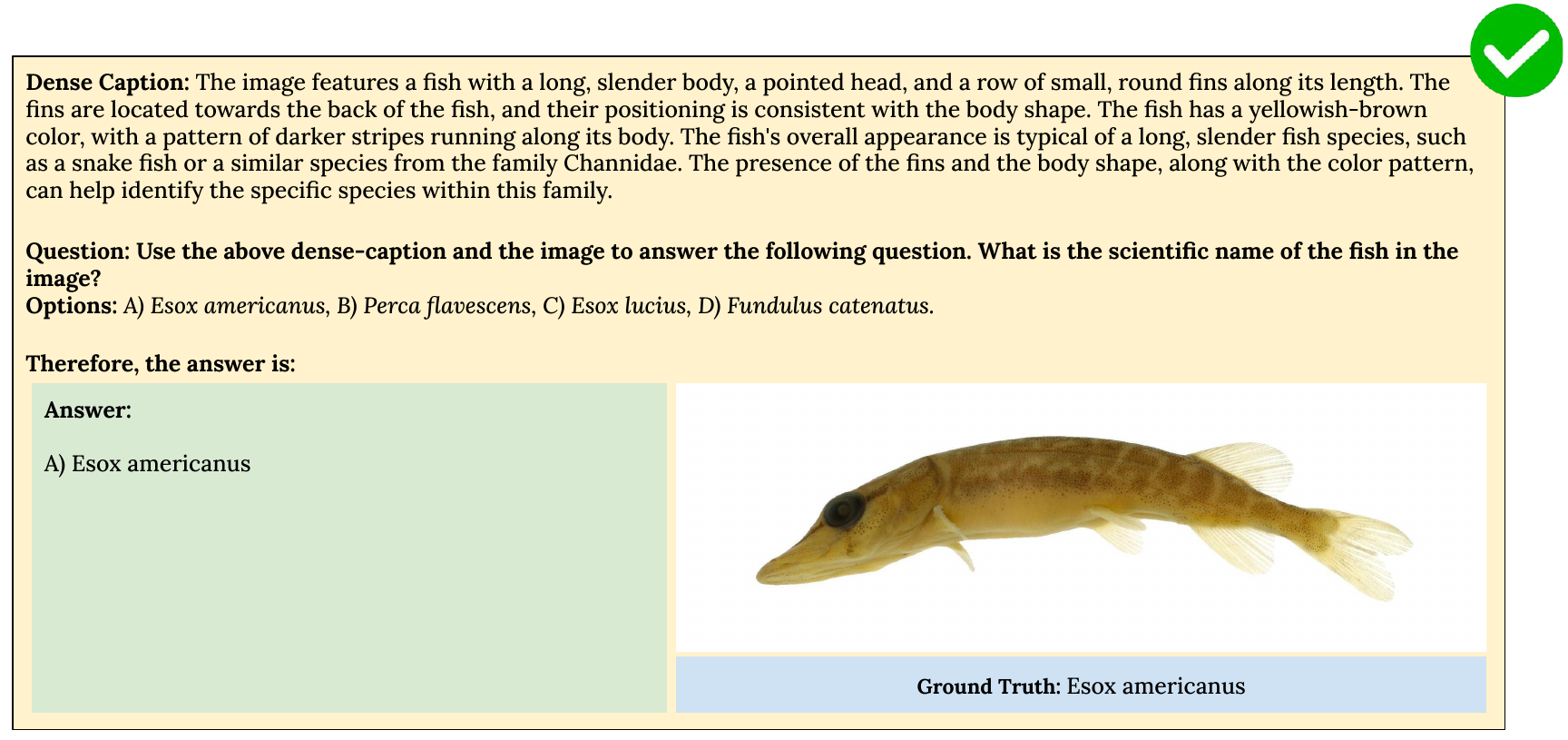}
    \caption{Dense Captions in Prompts. LLaVa-7B Correct prediction. Section \ref{sec:dense_caption}.}
    \label{llava7B_correct_dense_cap_fish}
\end{figure}

\begin{figure}[ht]
    \centering
    \includegraphics[width=\textwidth]{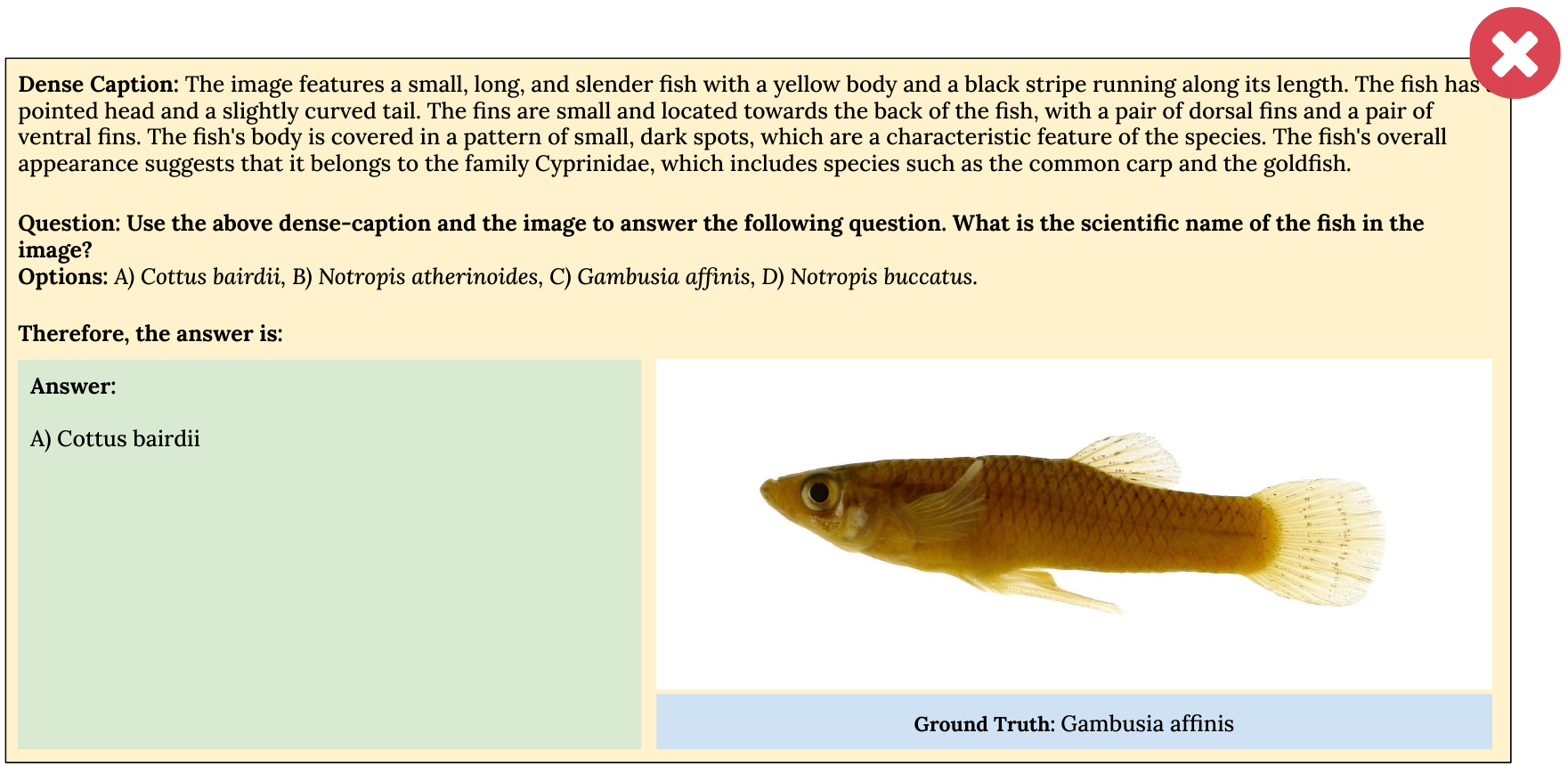}
    \caption{Dense Captions in Prompts. LLaVa-7B Incorrect prediction. Section \ref{sec:dense_caption}.}
    \label{llava7b_incorrect_dense_cap_fish}
\end{figure}

\begin{enumerate}
    \item Dense Captions in Prompts. GPT-4o Correct prediction. Refer to Figure \ref{gpt_correct_dense_cap_fish}.
    \item Dense Captions in Prompts. GPT-4o Incorrect prediction. Refer to Figure \ref{gpt_incorrect_dense_cap_fish}.
    \item Dense Captions in Prompts. LLaVa-7B Correct prediction. Refer to Figure \ref{llava7B_correct_dense_cap_fish}.
    \item Dense Captions in Prompts. LLaVa-7B Incorrect prediction. Refer to Figure \ref{llava7b_incorrect_dense_cap_fish}.
\end{enumerate}

\subsection{Chain-Of-Thought Prompting}
\label{sec:cot}
\begin{figure}[ht]
    \centering
    \includegraphics[width=\textwidth]{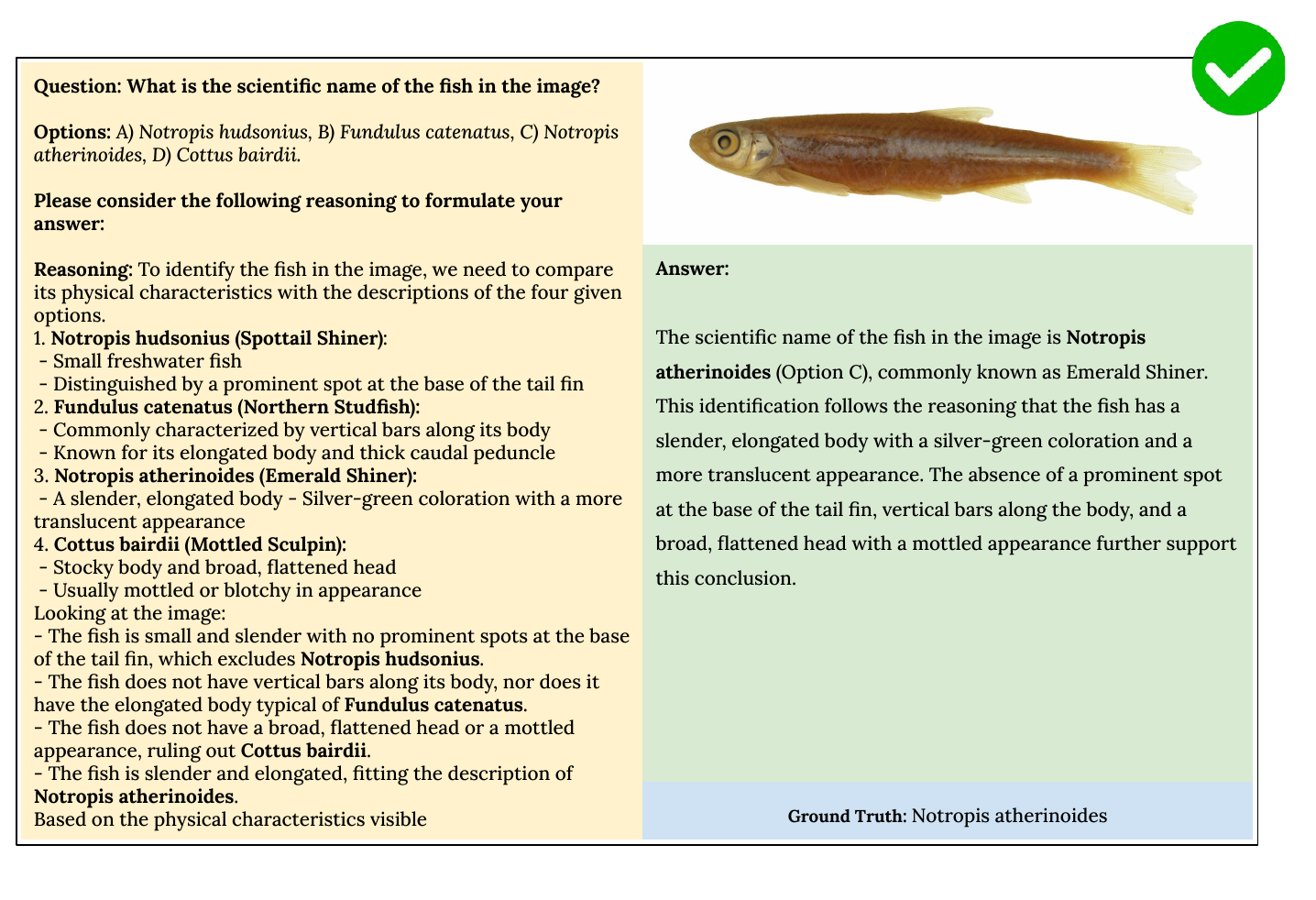}
    \caption{Chain-Of-Thought Prompting. GPT-4o Correct prediction. Section \ref{sec:cot}.}
    \label{gpt_correct_cot_fish}
\end{figure}

\begin{figure}[ht]
    \centering
    \includegraphics[width=\textwidth]{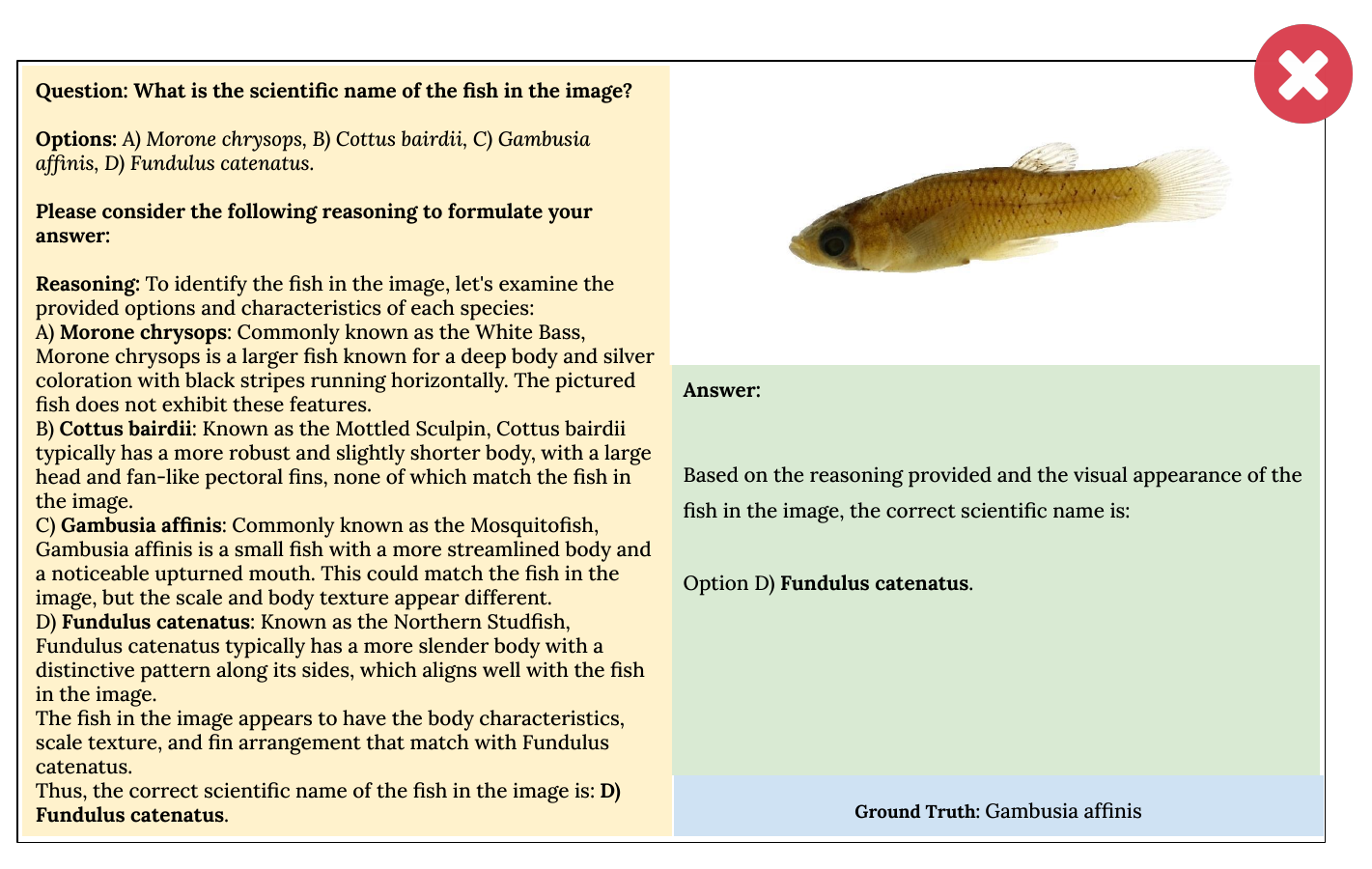}
    \caption{Chain-Of-Thought Prompting. GPT-4o Incorrect prediction. Section \ref{sec:cot}.}
    \label{gpt_incorrect_cot_fish}
\end{figure}

\begin{figure}[ht]
    \centering
    \includegraphics[width=\textwidth]{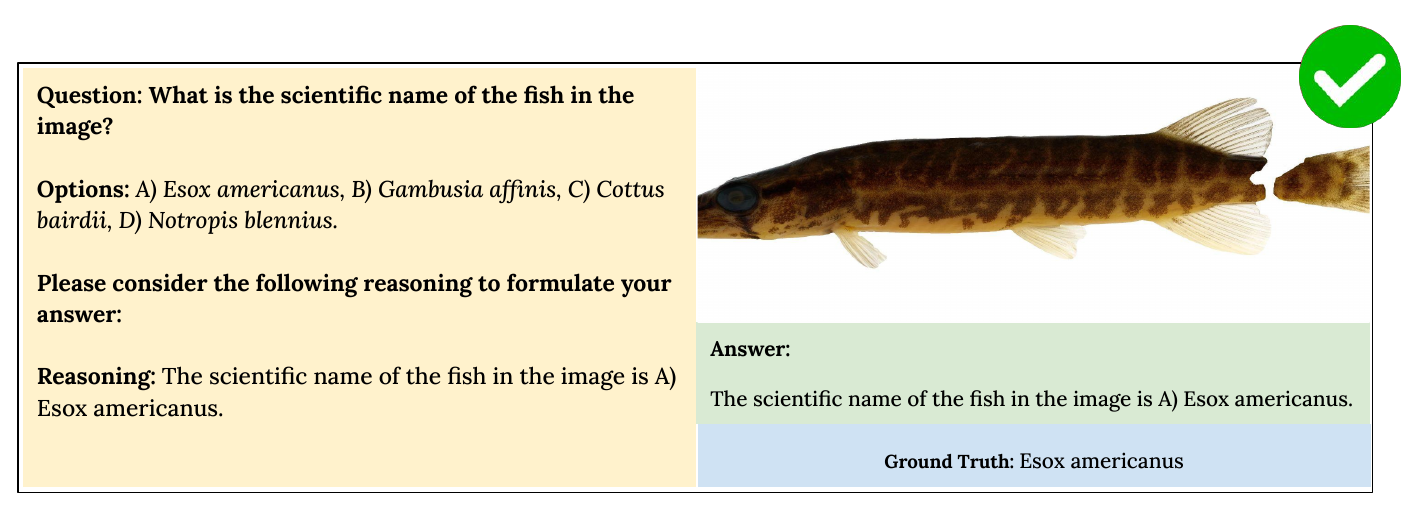}
    \caption{Chain-Of-Thought Prompting. LLaVa-13B Correct prediction. Section \ref{sec:cot}.}
    \label{llava13b_correct_cot_fish}
\end{figure}

\begin{figure}[ht]
    \centering
    \includegraphics[width=\textwidth]{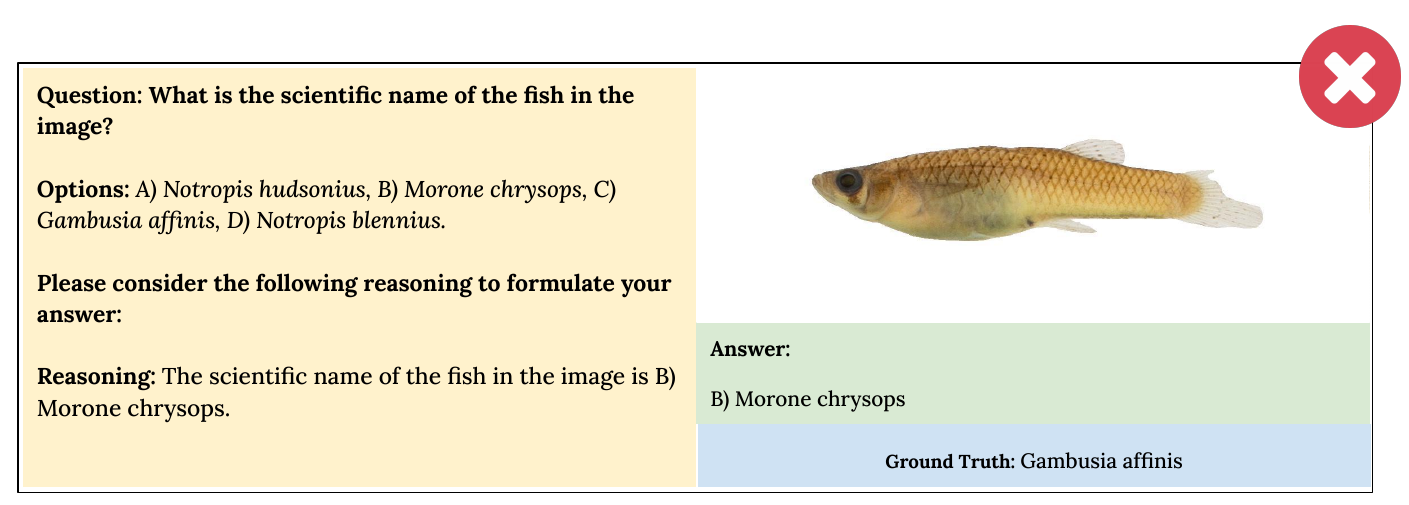}
    \caption{Chain-Of-Thought Prompting. LLaVa-13B Incorrect prediction. Section \ref{sec:cot}.}
    \label{llava13b_incorrect_cot_fish}
\end{figure}

\begin{enumerate}
    \item Chain-Of-Thought Prompting. GPT-4o Correct prediction. Refer to Figure \ref{gpt_correct_cot_fish}.
    \item Chain-Of-Thought Prompting. GPT-4o Incorrect prediction. Refer to Figure \ref{gpt_incorrect_cot_fish}.
    \item Chain-Of-Thought Prompting. LLaVa-13B Correct prediction. Refer to Figure \ref{llava13b_correct_cot_fish}.
    \item Chain-Of-Thought Prompting. LLaVa-13B Incorrect prediction. Refer to Figure \ref{llava13b_incorrect_cot_fish}.
\end{enumerate}

\section{Case Studies for Reasoning Hallucination Tests}

\subsection{False Confidence Test (FCT)}
\label{sec:fct}
\begin{figure}[t]
    \centering
    \includegraphics[width=\textwidth]{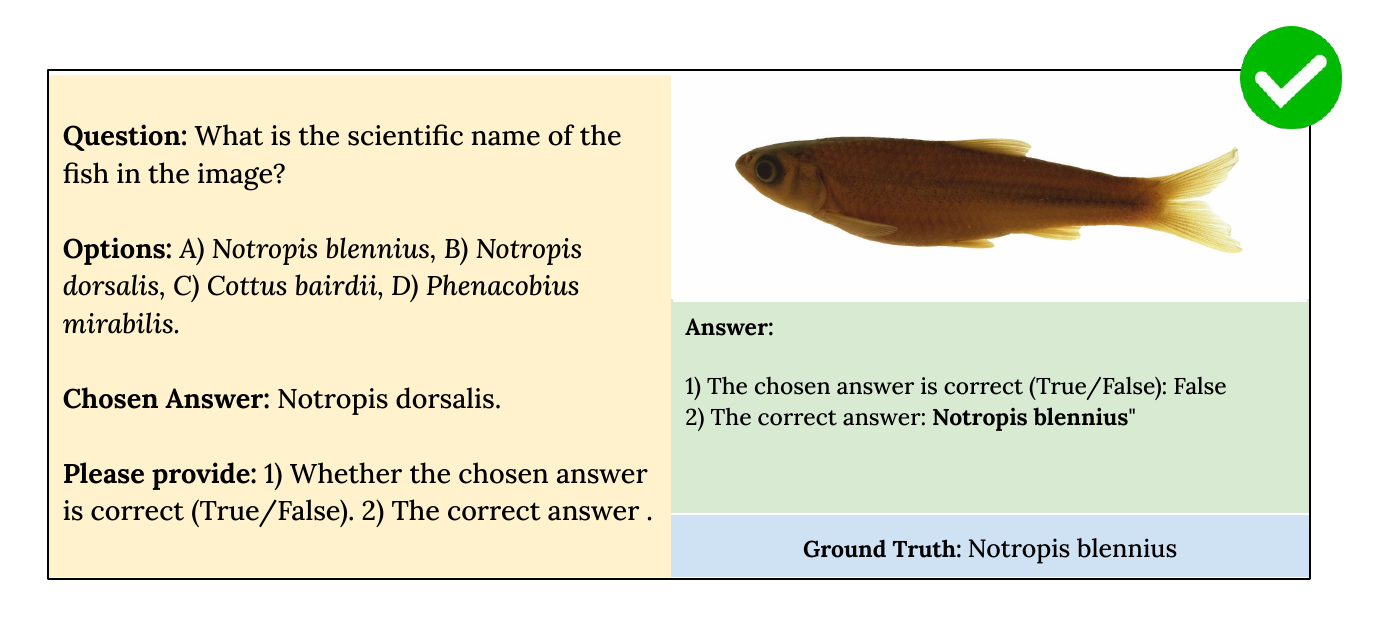}
    \caption{GPT-4o correct prediction. Section \ref{sec:fct}.}
    \label{gpt_correct_FCT_Fish_sample_3}
\end{figure}

\begin{figure}[t]
    \centering
    \includegraphics[width=\textwidth]{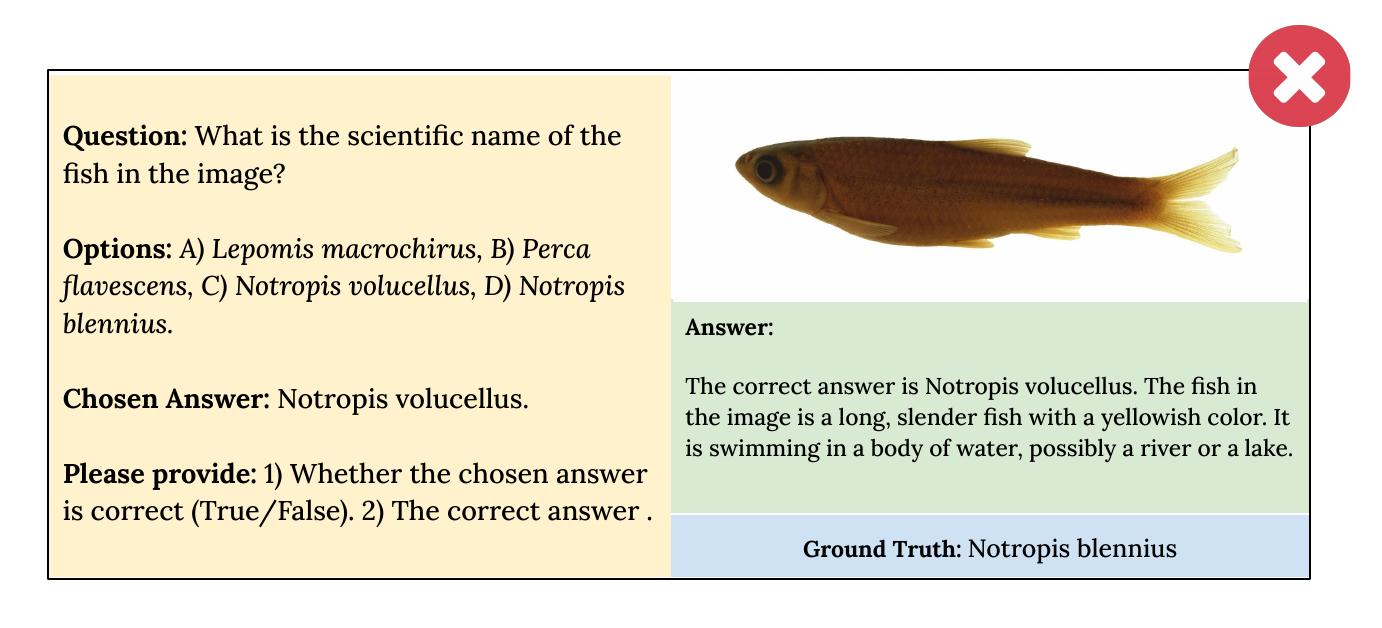}
    \caption{LLaVa-13B incorrect prediction. Section \ref{sec:fct}.}
    \label{llava_incorrect_FCT_Fish_sample_3}
\end{figure}

\begin{figure}[t]
    \centering
    \includegraphics[width=\textwidth, height=8.7cm]{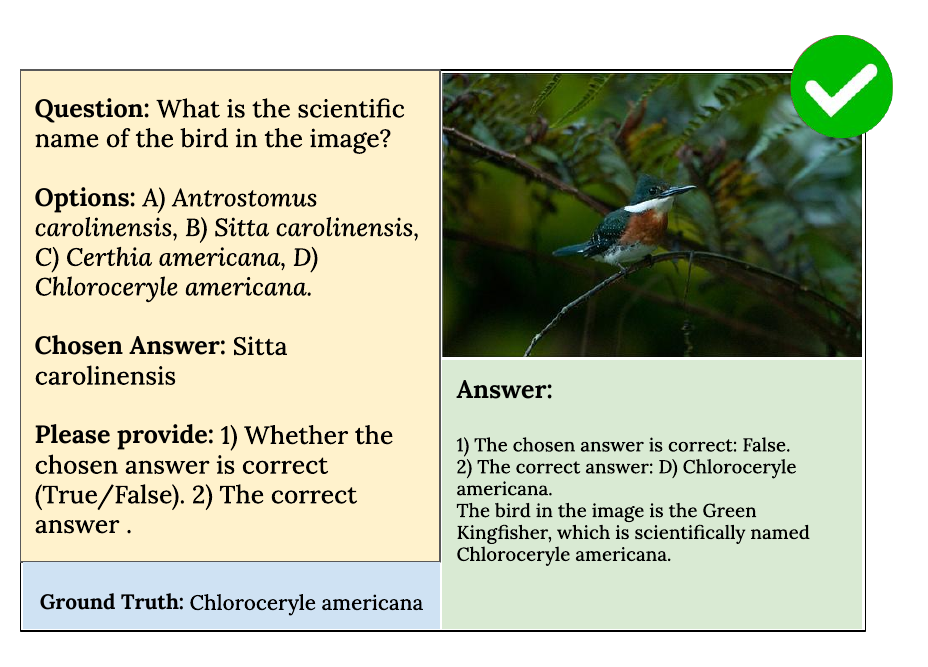}
    \caption{GPT-4o Correct prediction. Section \ref{sec:fct}.}
    \label{gpt_correct_FCT_Bird_sample_2}
\end{figure}

\begin{figure}[t]
    \centering
    \includegraphics[width=\textwidth, height=8.7cm]{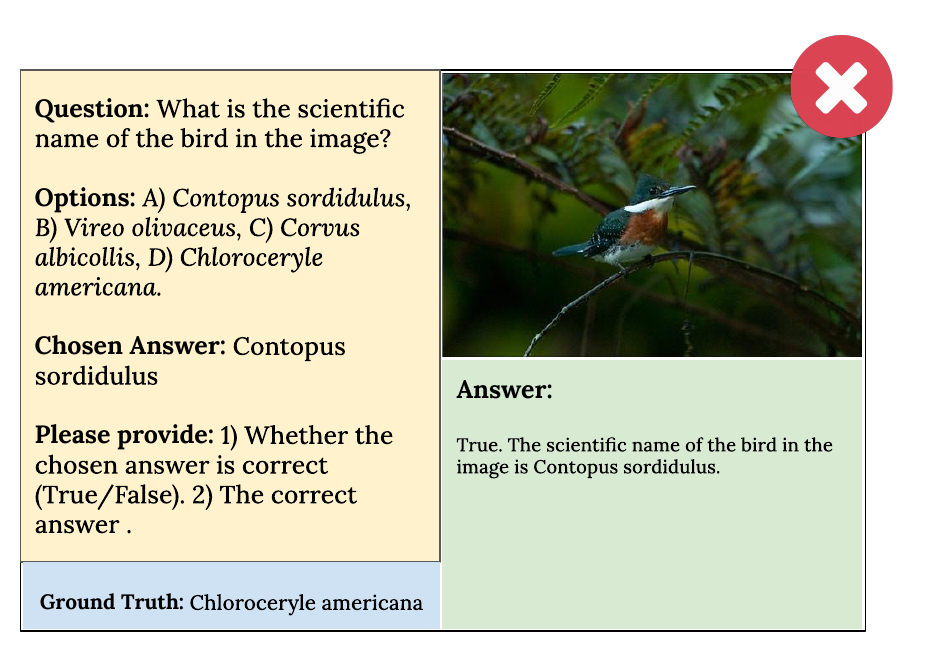}
    \caption{LLaVa-13B Incorrect prediction. Section \ref{sec:fct}.}
    \label{llava_incorrect_FCT_Bird_sample_2}
\end{figure}

\begin{figure}[t]
    \centering
    \includegraphics[width=\textwidth, height=8.7cm]{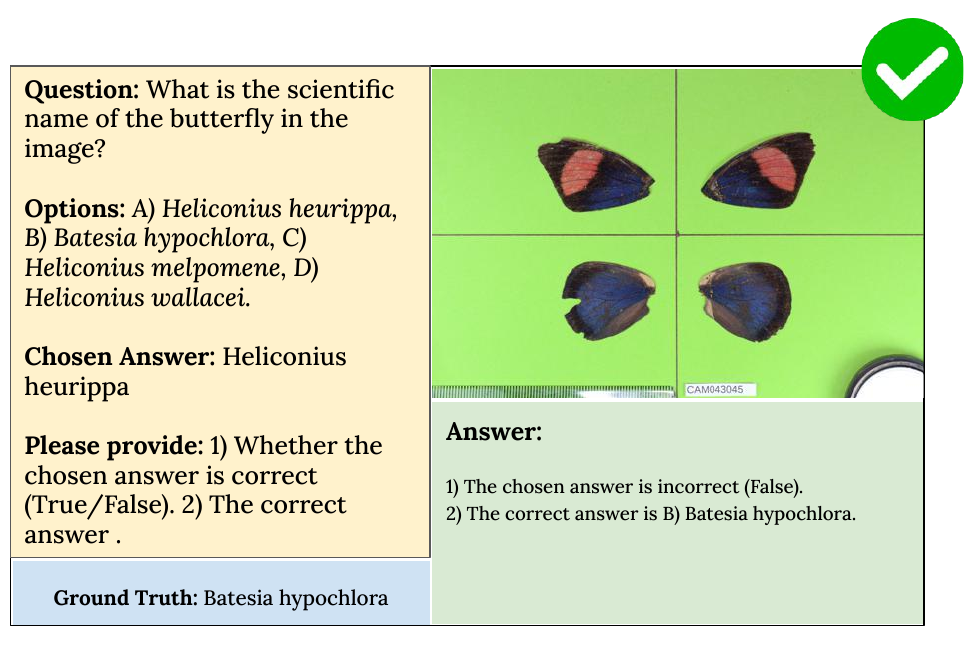}
    \caption{GPT-4o Correct prediction. Section \ref{sec:fct}.}
    \label{gpt_correct_FCT_Butterfly_sample_3}
\end{figure}

\begin{figure}[t]
    \centering
    \includegraphics[width=\textwidth, height=8.7cm]{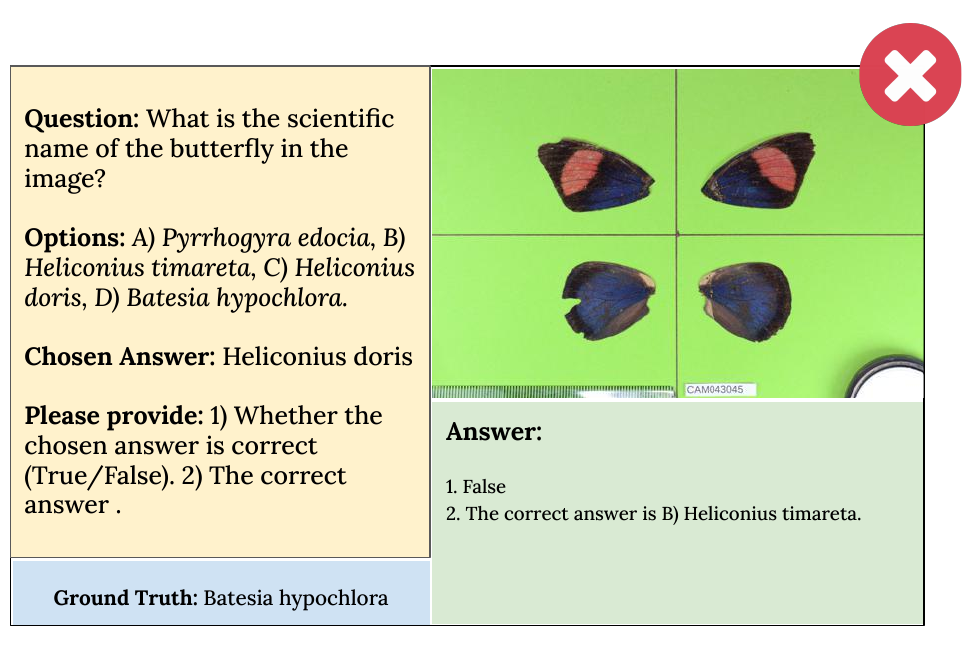}
    \caption{LLaVa-13B Incorrect prediction. Section \ref{sec:fct}.}
    \label{llava_incorrect_FCT_Butterfly_sample_3}
\end{figure}

\begin{enumerate}
    \item FCT test on Fish dataset. GPT-4o Correct prediction. Refer to Figure \ref{gpt_correct_FCT_Fish_sample_3}.
    \item FCT test on Fish dataset. LLaVa-13B Incorrect prediction. Refer to Figure \ref{llava_incorrect_FCT_Fish_sample_3}.
    \item FCT test on Bird dataset. GPT-4o Correct prediction. Refer to Figure \ref{gpt_correct_FCT_Bird_sample_2}.
    \item FCT test on Bird dataset. LLaVa-13B Incorrect prediction. Refer to Figure \ref{llava_incorrect_FCT_Bird_sample_2}.
    \item FCT test on Butterfly dataset. GPT-4o Correct prediction. Refer to Figure \ref{gpt_correct_FCT_Butterfly_sample_3}.
    \item FCT test on Butterfly dataset. LLaVa-13B Incorrect prediction. Refer to Figure \ref{llava_incorrect_FCT_Butterfly_sample_3}.
\end{enumerate}

\subsection{None of The Above (NOTA) Test}

\label{sec:nota}
\begin{figure}[t]
    \centering
    \includegraphics[width=\textwidth]{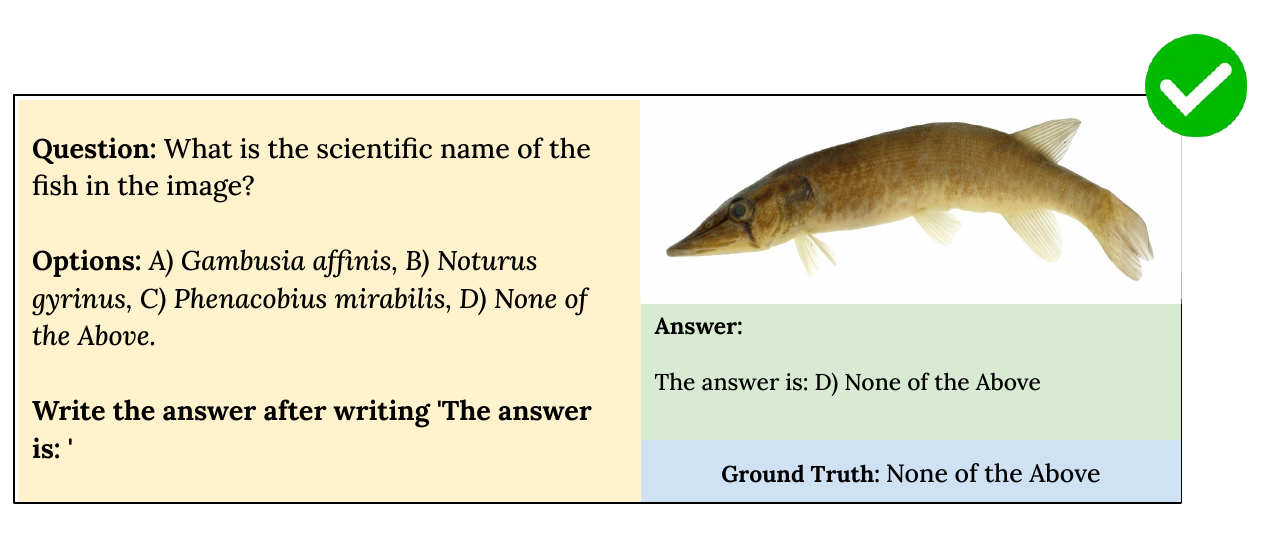}
    \caption{GPT-4o Correct prediction. Actual species name is Esox Americanus. Section \ref{sec:nota}.}
    \label{gpt_correct_esox_americanus_NOTA_Fish_sample_3}
\end{figure}

\begin{figure}[t]
    \centering
    \includegraphics[width=\textwidth]{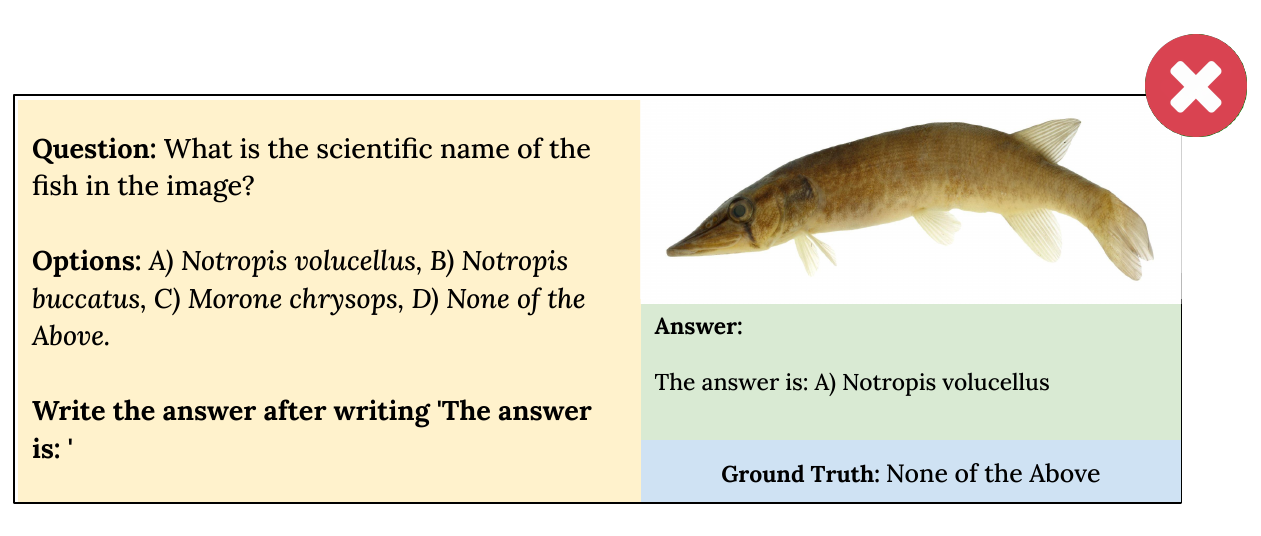}
    \caption{LLaVa-13B Incorrect prediction. Actual species name is Esox Americanus. Section \ref{sec:nota}.}
    \label{llava_incorrect_esox_americanus_NOTA_Fish_sample_3}
\end{figure}

\begin{figure}[t]
    \centering
    \includegraphics[width=\textwidth, height=7.5cm]{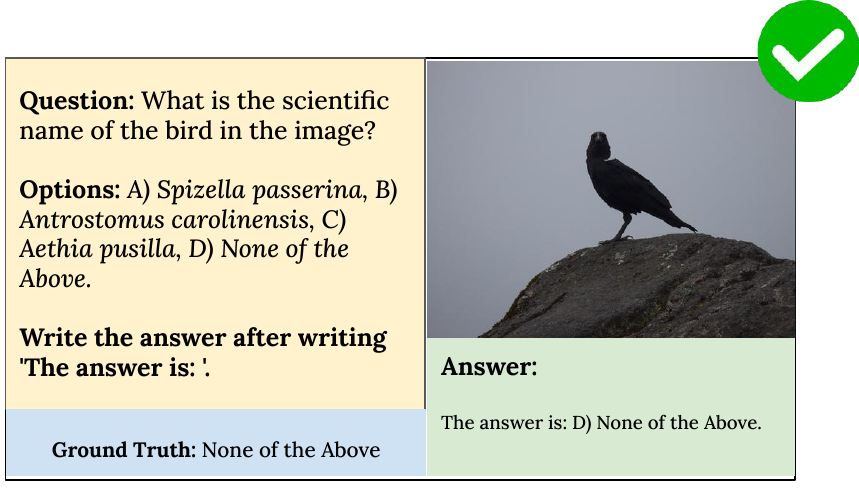}
    \caption{GPT-4o Correct prediction. Actual species name is Corvus Albicollis. Section \ref{sec:nota}.}
    \label{gpt_correct_corvus_albicollis_NOTA_Bird_sample_3}
\end{figure}

\begin{figure}[t]
    \centering
    \includegraphics[width=\textwidth, height=7.5cm]{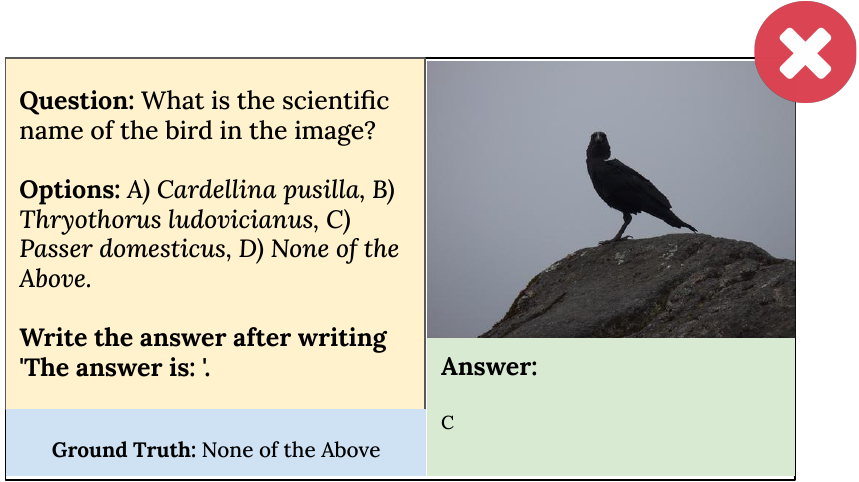}
    \caption{Blip-Flan-XL Incorrect prediction. Actual species name is Corvus Albicollis. Section \ref{sec:nota}.}
    \label{blip_flan_xl_incorrect_corvus_albicollis_NOTA_Bird_sample_3}
\end{figure}

\begin{figure}[t]
    \centering
    \includegraphics[width=\textwidth, height=7.5cm]{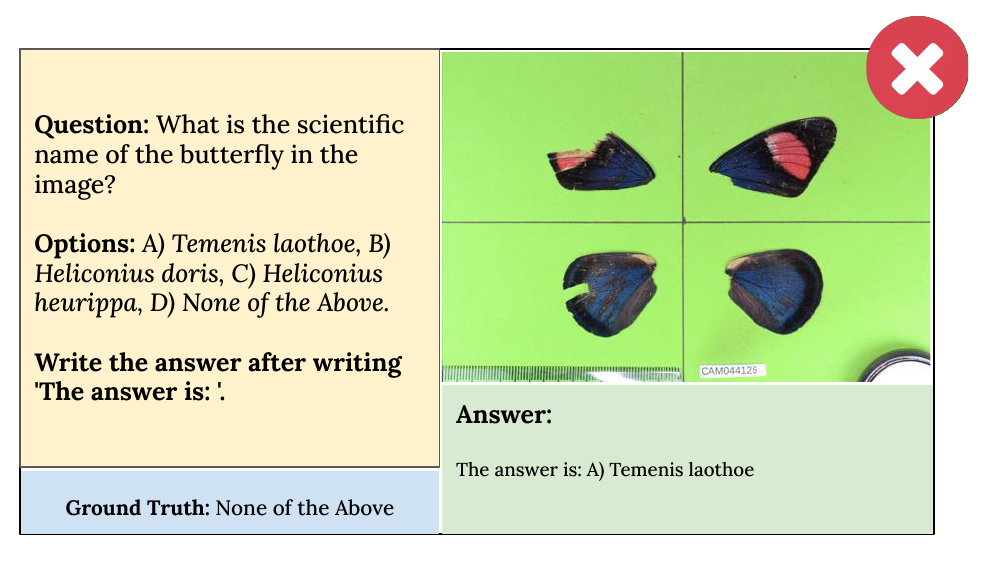}
    \caption{GPT-4o Incorrect prediction. Actual species name is \textit{Batesia Hypochlora}. Section \ref{sec:nota}.}
    \label{gpt_incorrect_batesia_hypochlora_NOTA_Butterfly_sample_3}
\end{figure}

\begin{figure}[t]
    \centering
    \includegraphics[width=\textwidth, height=7.5cm]{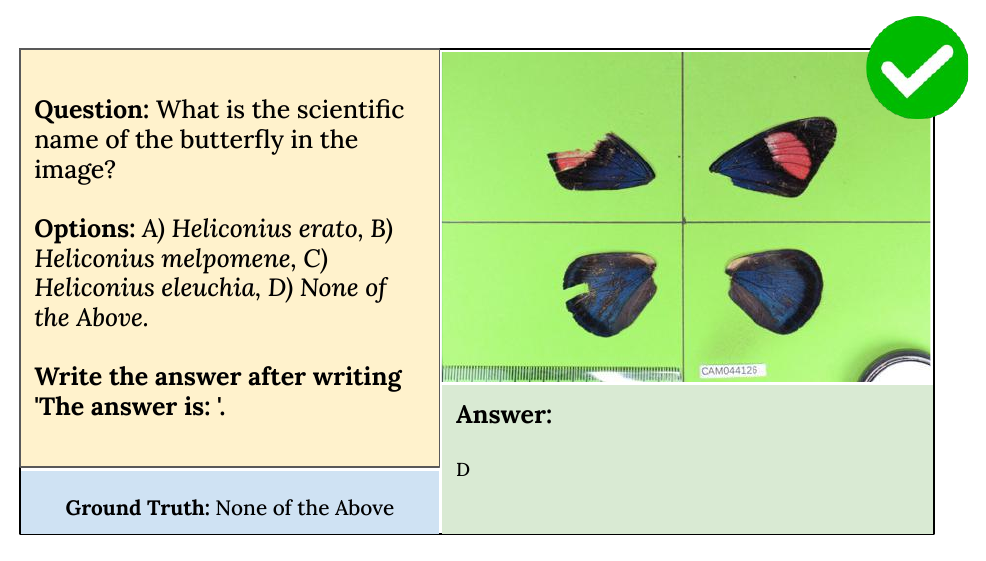}
    \caption{Blip-Flan-XL Correct prediction. Actual species name is \textit{Batesia Hypochlora}. Section \ref{sec:nota}.}
    \label{blip_flan_xl_correct_batesia_hypochlora_NOTA_Butterfly_sample_3}
\end{figure}

\begin{enumerate}
    \item NOTA test on Fish dataset. GPT-4o Correct prediction. Actual species name is \textit{Esox Americanus}. Refer to Figure \ref{gpt_correct_esox_americanus_NOTA_Fish_sample_3}.
    \item NOTA test on Fish dataset. LLaVa-13B Incorrect prediction. Actual species name is \textit{Esox Americanus}. Refer to Figure \ref{llava_incorrect_esox_americanus_NOTA_Fish_sample_3}.
    \item NOTA test on Bird dataset. GPT-4o Correct prediction. Actual species name is \textit{Corvus Albicollis}. Refer to Figure \ref{gpt_correct_corvus_albicollis_NOTA_Bird_sample_3}.
    \item NOTA test on Bird dataset. Blip-Flan-XL Incorrect prediction. Actual species name is \textit{Corvus Albicollis}. Refer to Figure \ref{blip_flan_xl_incorrect_corvus_albicollis_NOTA_Bird_sample_3}.
    \item NOTA test on Butterfly dataset. GPT-4o Incorrect prediction. Actual species name is \textit{Batesia Hypochlora}. Refer to Figure \ref{gpt_incorrect_batesia_hypochlora_NOTA_Butterfly_sample_3}.
    \item NOTA test on Butterfly dataset. Blip-Flan-XL Correct prediction. Actual species name is \textit{Batesia Hypochlora}. Refer to Figure \ref{blip_flan_xl_correct_batesia_hypochlora_NOTA_Butterfly_sample_3}.
\end{enumerate}

\clearpage































\end{document}